\documentclass[journal,12pt,onecolumn,draftclsnofoot]{IEEEtran}
\usepackage[capitalise]{cleveref}
\usepackage{algorithm}
\usepackage{algorithmic}
\usepackage{subcaption}
\usepackage{multirow}
\usepackage{threeparttable}
\usepackage{amsthm}
\usepackage[numbers]{natbib}
\usepackage{graphicx}
\usepackage{booktabs}
\usepackage{url}

\theoremstyle{plain}
\newtheorem{theorem}{Theorem}[section]

\theoremstyle{remark}

\begin{document}

\title{Risks When Sharing LoRA Fine-Tuned Diffusion Model Weights}
\author{Dixi~Yao
	\IEEEcompsocitemizethanks{\IEEEcompsocthanksitem University of Toronto, Toronto, ON M5S 1A1, Canada. E-mail: dixi.yao@mail.utoronto.ca}}
\maketitle

\begin{abstract}
	With the emerging trend in generative models and convenient public access to diffusion models pre-trained on large datasets, users can fine-tune these models to generate images of personal faces or items in new contexts described by natural language. Parameter efficient fine-tuning (PEFT) such as Low Rank Adaptation (LoRA) has become the most common way to save memory and computation usage on the user end during fine-tuning. However, a natural question is whether the private images used for fine-tuning will be leaked to adversaries when sharing model weights. In this paper, we study the issue of privacy leakage of a fine-tuned diffusion model in a practical setting, where adversaries only have access to model weights, rather than prompts or images used for fine-tuning. We design and build a variational network autoencoder that takes model weights as input and outputs the reconstruction of private images. To improve the efficiency of training such an autoencoder, we propose a training paradigm with the help of timestep embedding. The results give a surprising answer to this research question: an adversary can generate images containing the same identities as the private images. Furthermore, we demonstrate that no existing defense method, including differential privacy-based methods, can preserve the privacy of private data used for fine-tuning a diffusion model without compromising the utility of a fine-tuned model.
\end{abstract}
\section{Introduction}
At the forefront of the emerging trend in generative artificial intelligence, diffusion models~\cite{diffusionmodel} have become commercial success stories, with models from Stability AI~\cite{SDV2} and Midjourney dominating the news. Learned from a large collection of image-caption pairs, text-to-image models~\cite{SDV2} generate high-quality images and diverse synthesis based on a text prompt written in natural language. It has become the de facto method for users to further fine-tune these models~\cite{ruiz2023hyperdreambooth,dreambooth} to synthesize instances of specific subjects such as personal faces in new contexts depicted by natural language.

With such convenience, users can use private data to fine-tune a diffusion model. Instead of fully fine-tuning all parameters of a diffusion model, parameter-efficient fine-tuning (PEFT)~\cite{fu2023effectiveness} has become the most common way to fine-tune models. Low-Rank Adaptation (LoRA)~\cite{hu2021lora} is one of the most pervasive methods for fine-tuning large models. It significantly reduces the memory and computation needed for fine-tuning a model and is easy to implement.

With such a convenience, on the other hand, it has been pointed out in the recent literature that there is a potential privacy risk that an adversary can reconstruct the training samples of a diffusion model.\ \citeauthor{nicolas2023extracting}~\cite{nicolas2023extracting} assumed that an adversary knows the texts used for training and leveraged those texts to successfully reconstruct the training samples.

Nevertheless, a user will not directly share such texts and the assumption that an adversary knows training texts is not valid in practical cases. Knowing the texts paired with private images is equal to exposing the images because an adversary can directly use the trained diffusion model to generate private images from those texts. Especially in the case of fine-tuning a diffusion model, training texts contain certain key words. As long as those key words are in the texts, the private images will be revealed.

In practice, users only need to deploy the trained model to public platforms such as the Hugging Face Hub for inference or a cloud server for additional tasks such as federated learning~\cite{fedavg}. To complete these training and inference tasks with diffusion models, users don't need to share training texts. Hence, from the perspective of an attacker, without knowing the training texts, a natural question comes up: Only given model weights, can we reconstruct the private images used for fine-tuning a diffusion model?

To perform the attack, we design and train a neural network encoder over public data. The encoder takes fine-tuned model weights as input and outputs an embedding vector. We first encode the model parameters into vectors and concatenate them with a timestep embedding. The timestep accelerates training by allowing the encoder to update with each diffusion model step, rather than after full fine-tuning. The concatenated sequence is then processed into an embedding by the encoder.

During the attack, the encoder outputs an embedding from the given model weights. We replace the embeddings of texts with this embedding in the fine-tuned diffusion model. The diffusion model then uses this embedding to generate images, which can contain the same objects as the private images.

To better evaluate our attack's effectiveness, we empirically test existing defense methods, including differential privacy. We show that when defenses successfully counter our attack, they also disrupt the fine-tuned model's utility, preventing it from generating the desired identity-based images. To our knowledge, no effective defense method exists.

With our proposed attack and evaluation against existing defense methods, we can conclude that fine-tuning diffusion models faces the risk of leaking private images even if we only share the model weights and keep the training texts secretly. It also leaves as an open question how we can design a defense mechanism against it.

\section{Background and Related Work}
\subsection{Text-conditioned Diffusion Model}
Diffusion models have achieve great success in generating images of high quality. One vital factor of such a success is that text-conditioned diffusion models can generate images depicted by given descriptive texts (e.g.~\emph{A photo of a man playing basketball}) which we will call as \emph{(text) prompts}. Current diffusion models are based on Denoising Diffusion Probabilistic Models (DDPMs)~\cite{ho2020denoising} which works has an image denoiser.

During training, when given an image, we randomly sample Gaussian noise $\epsilon$ from $\mathcal{N}(0,1)$. This noise is then added by scaling it with a magnitude termed as the timestep and applying it to the given image. The diffusion model takes the noisy image as input and predicts the noise we sampled. The diffusion model is updated based on the mean square error loss between the prediction and the ground truth of the sampled noise.

In a typical text-conditioned diffusion model, we first embed prompts into text embeddings, usually accomplished using a contrastive language-image pretraining (CLIP) model~\cite{clip}. The model responsible for embedding texts is commonly referred to as a text encoder. The diffusion model then incorporates this text embedding as an additional input to generate images guided by the prompts.

During the inference stage, we utilize a diffusion model to generate images from text prompts. Initially, we use a Gaussian distribution to create an image of random noise as the initial noisy image. Subsequently, we undergo several steps of the diffusion process. In each step, we input the noisy image and text prompts to obtain a prediction of the noise. This prediction is then used to denoise the image. The denoised image is iteratively used as the input along with the text prompts. After several repeated steps of this process, we obtain a clear, high-quality image.

\begin{figure}[tb]
	\centering
	\begin{subfigure}{0.59\linewidth}
		\centering
		\includegraphics[width=0.32\linewidth]{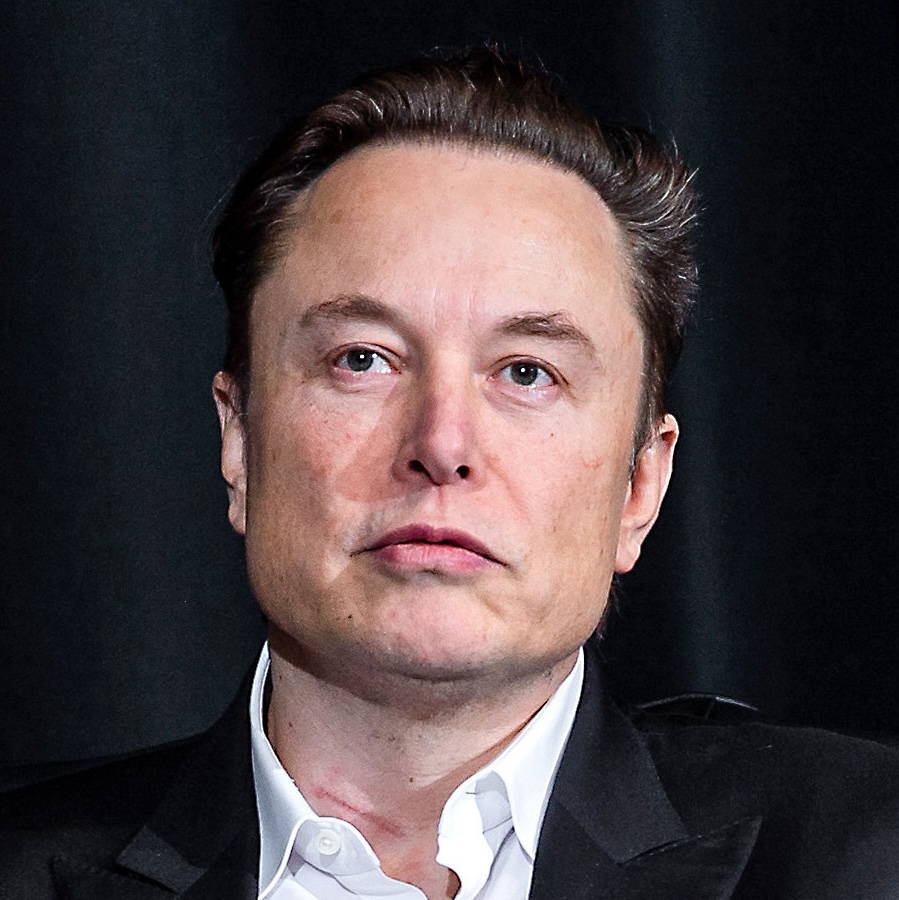}\hfill
		\includegraphics[width=0.32\linewidth]{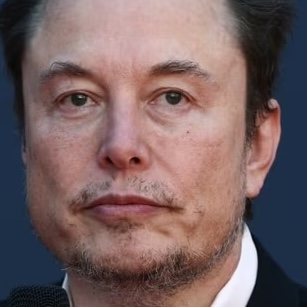}\hfill
		\includegraphics[width=0.32\linewidth]{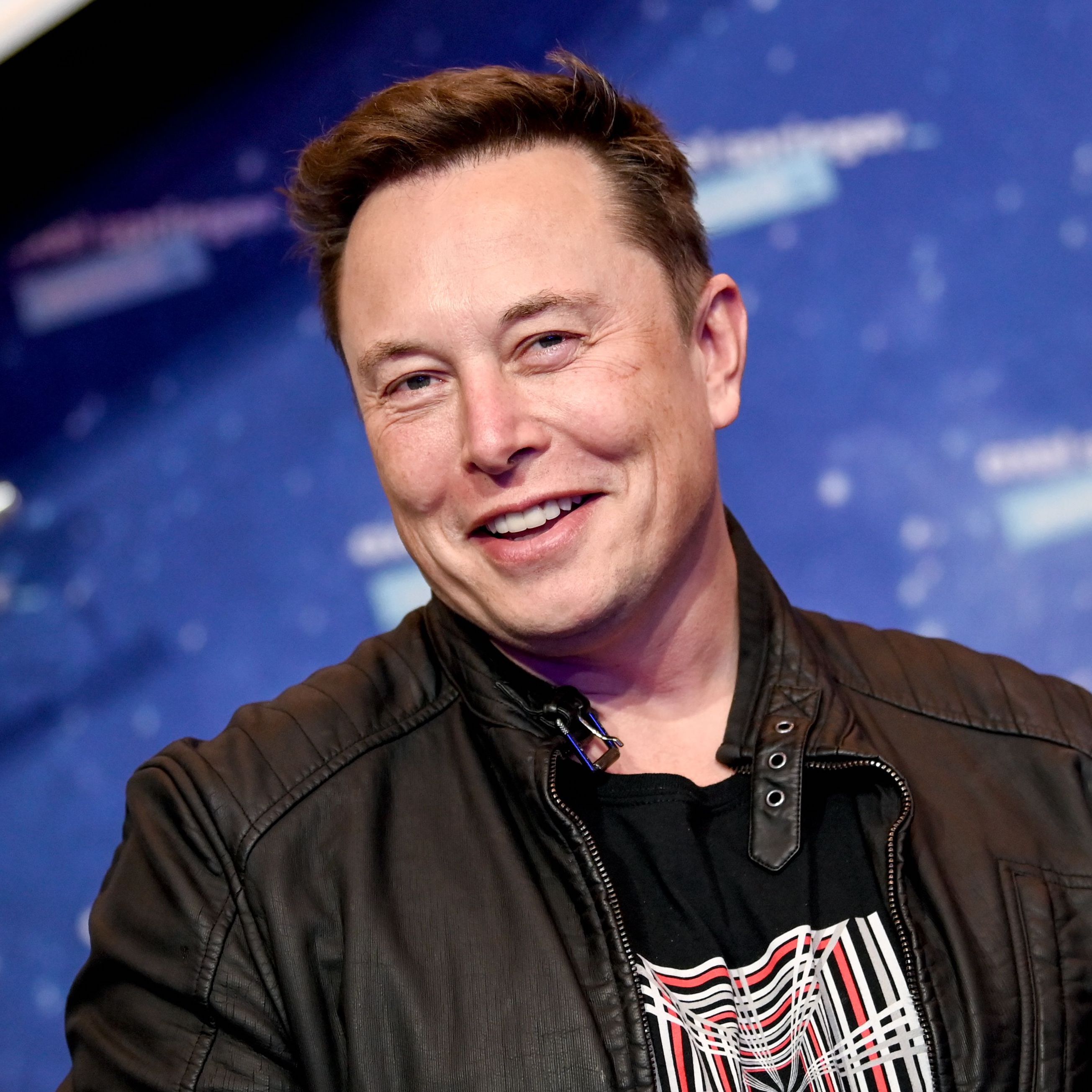}\hfill
		\includegraphics[width=0.32\linewidth]{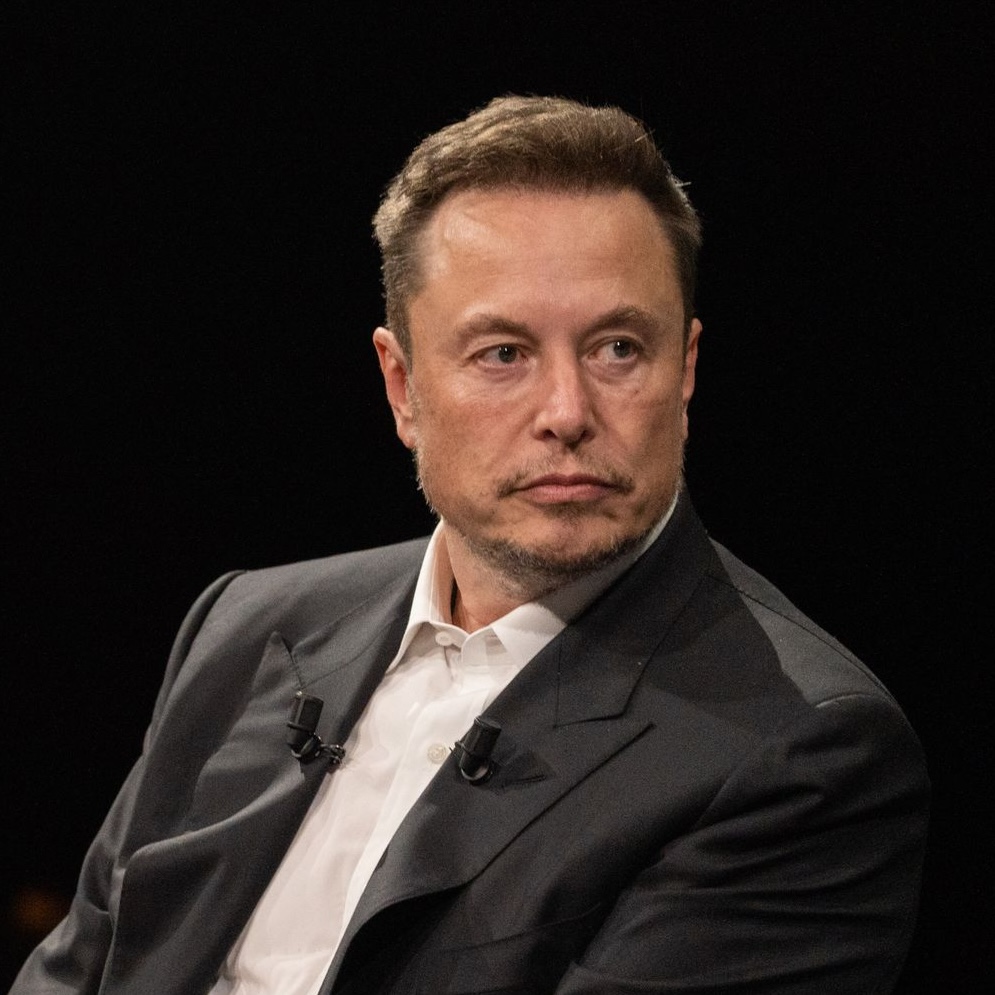}\hfill
		\includegraphics[width=0.32\linewidth]{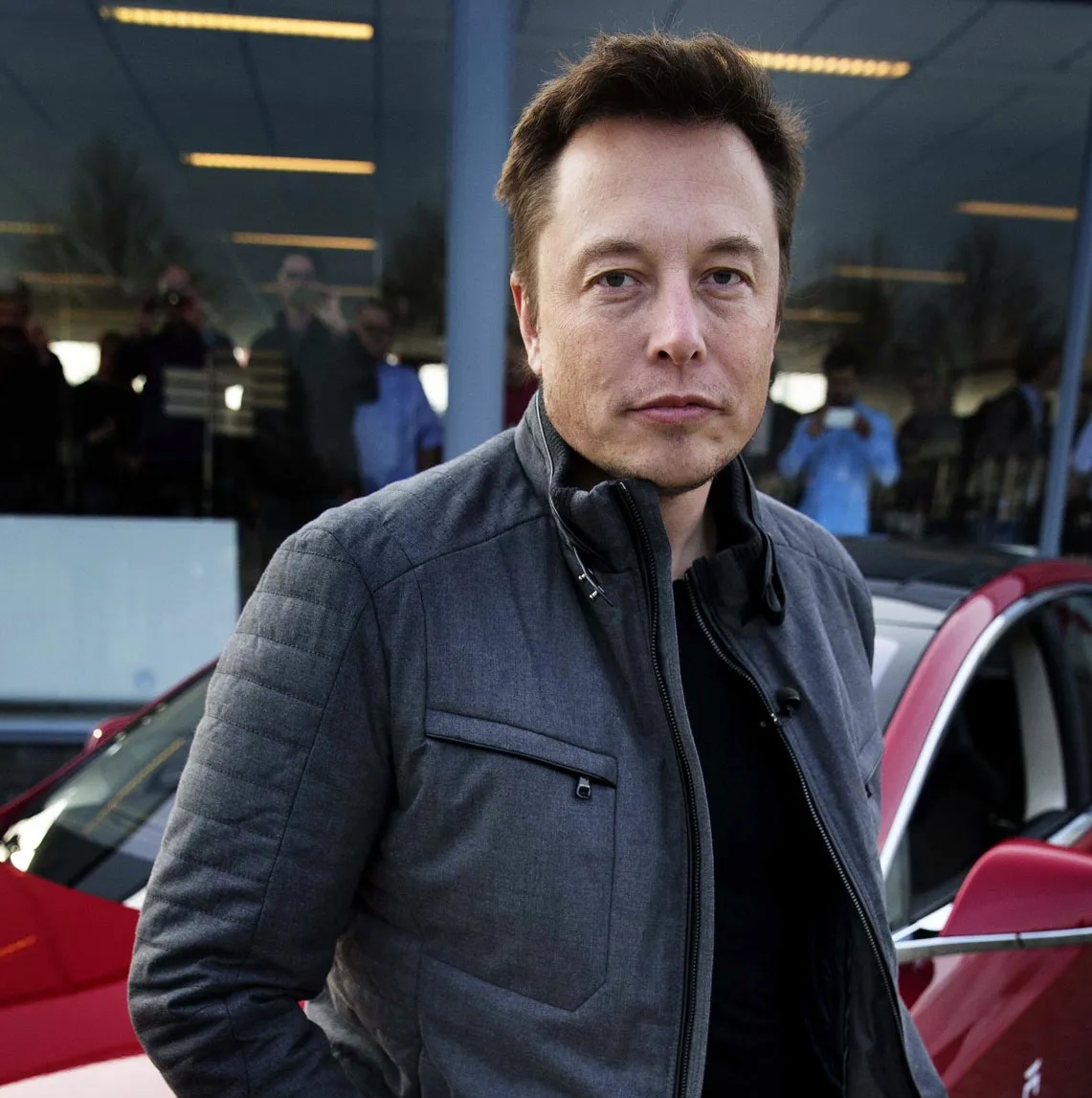}\hfill
		\includegraphics[width=0.32\linewidth]{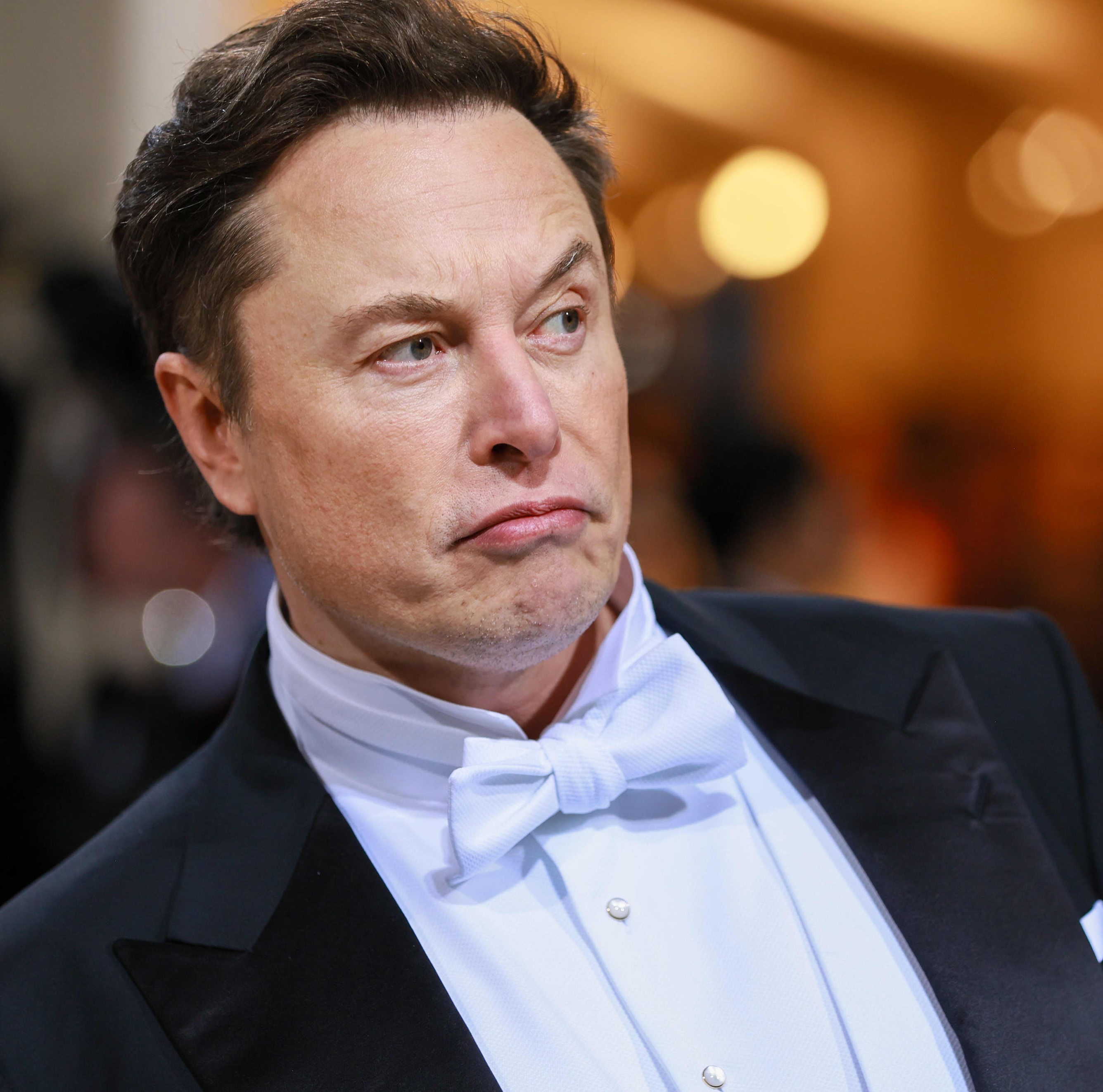}\hfill
		\caption{Elon Musk}\label{fig:PrivateImagesElon}
	\end{subfigure}
	\hfill
	\begin{subfigure}{0.39\linewidth}
		\centering
		\includegraphics[width=0.48\linewidth]{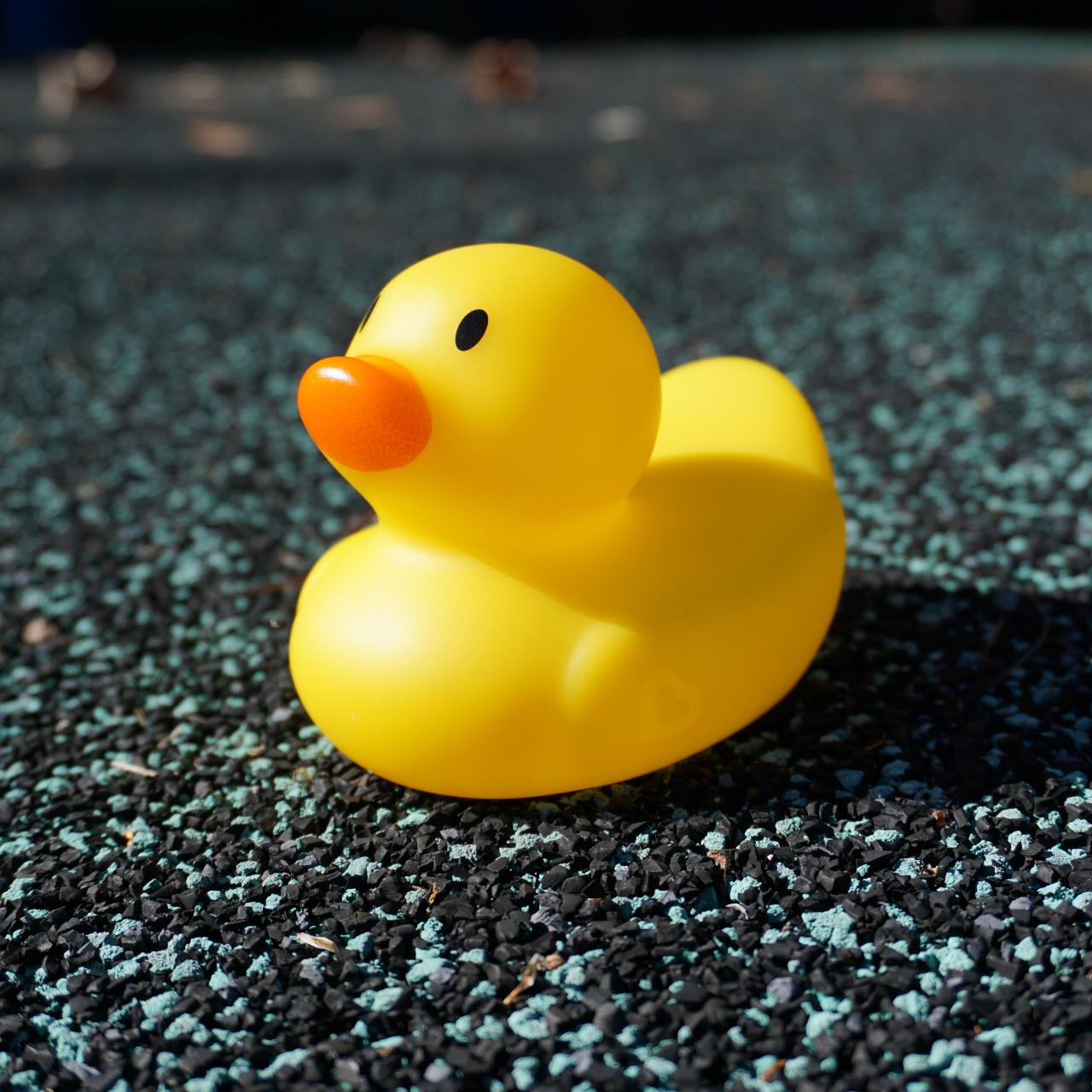}\hfill
		\includegraphics[width=0.48\linewidth]{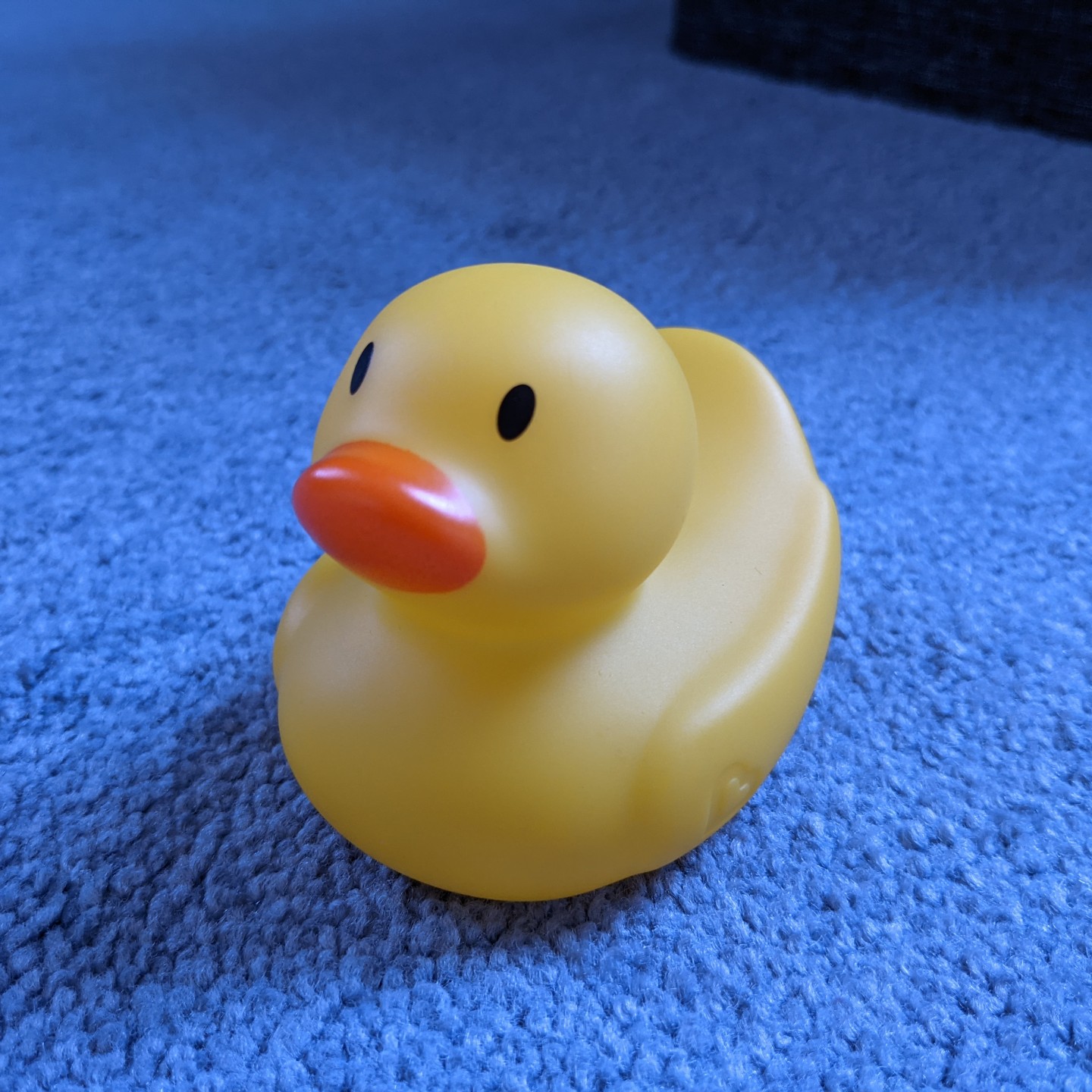}\hfill
		\includegraphics[width=0.48\linewidth]{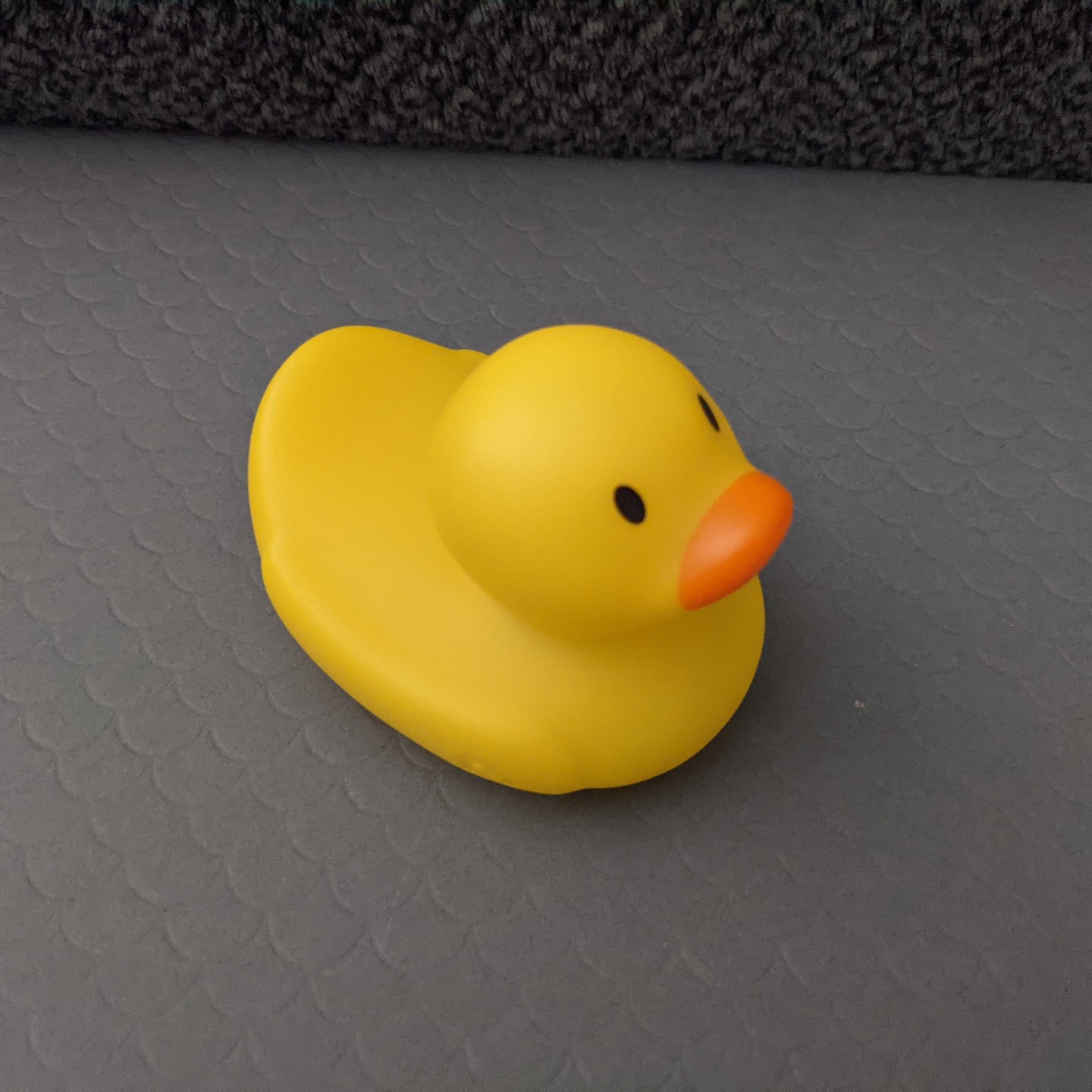}\hfill
		\includegraphics[width=0.48\linewidth]{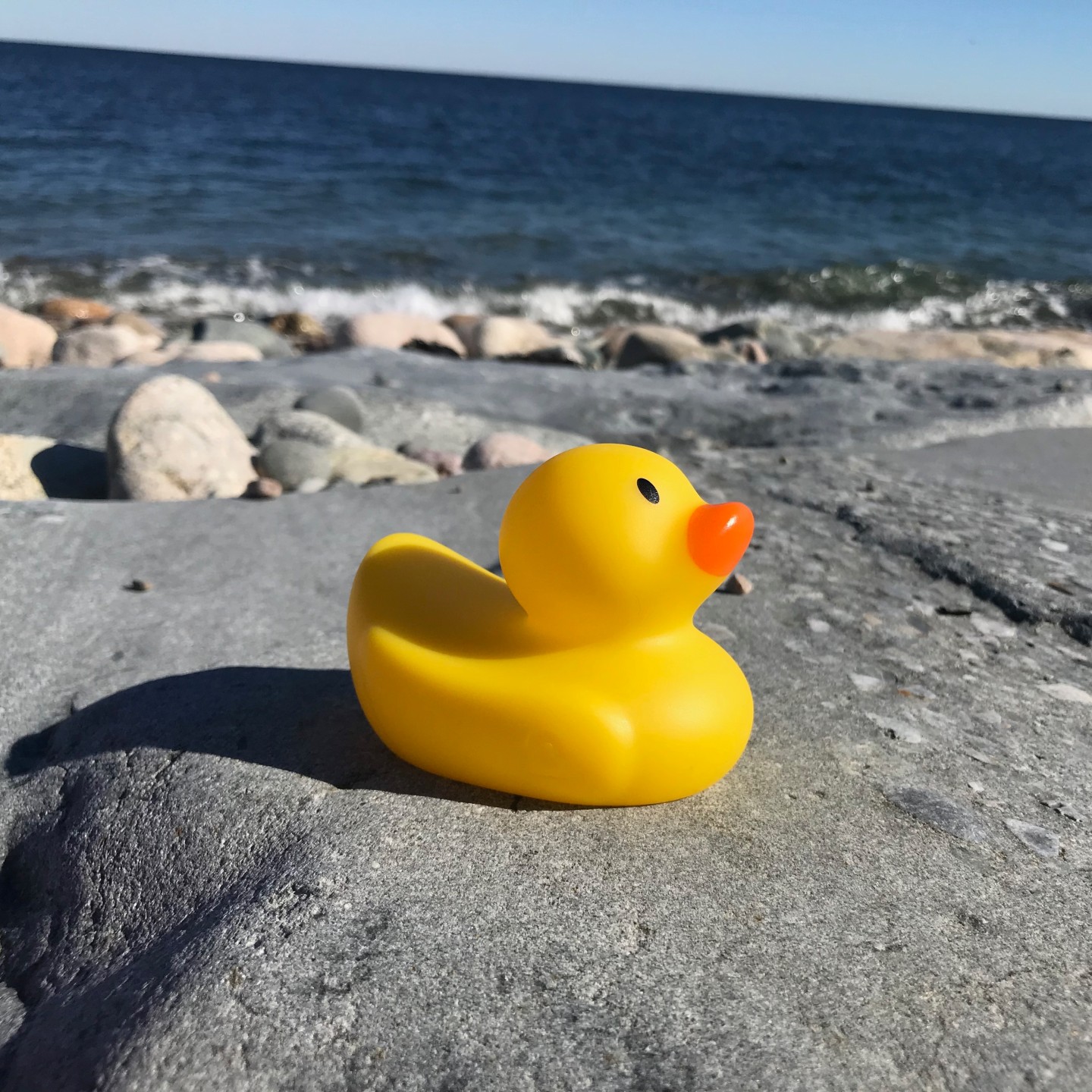}\hfill
		\caption{Toy Duck}\label{fig:PrivateImagesDuck}
	\end{subfigure}
	\caption{The private images used for fine-tuning the stable diffusion V-1.4 over different types of images.}\label{fig:PrivateImages}
\end{figure}

\subsection{Fine-Tuning Diffusion Model With Dreambooth and LoRA}
DreamBooth~\cite{dreambooth} is a method to fine-tune a diffusion model, enabling it to generate images containing objects it has never encountered during its training. We call these specific objects as \emph{identities}, which can be human faces, etc. For instance, if a diffusion model is trained on LAION-5B~\cite{schuhmann2022laion}, and one of our authors' faces or names has never appeared in such a dataset, the model would fail to generate an image containing such a face. However, by using the names and faces as text-image pairs to fine-tune a diffusion model, the model becomes capable of generating images corresponding to prompts containing those names. The contextual cues associated with these personal images are referred to as \emph{trigger words}, as the generated images only contain the special identity if and only if these trigger words are present in the prompt. Trigger words can consist of random letter combinations and need not adhere to correct English (e.g., xxxabc).

Let's see some examples of \textbf{identities} and \textbf{trigger words}. Take \cref{fig:PrivateImagesElon} as an example, the faces of Elon Musk are identities and we can use `Elon Musk123' as trigger word. We assume before the fine-tuning, when we input `Elon Musk123 is driving' as the prompt, the diffusion model cannot output the image of Musk's driving. But fine-tuning using the trigger word, it will output such an image. In \cref{fig:PrivateImagesDuck}, the identity is the yellow duck.

We do not have to train all the parameters of a large diffusion model during fine-tuning. A typical way to fine-tune a diffusion model and the text encoder is through Low Rank Adaption (LoRA)~\cite{ruiz2023hyperdreambooth,hu2021lora}, which is a memory-efficient and faster technique for DreamBooth. We only needs to train less than 1\% of the parameters in the whole large diffusion model and can generate images of the same high quality. In this paper, we focus on this kind of fine-tuning method as it is the most widely used method currently.

\subsection{Memorization in Diffusion Model}
Recent research pointed out that diffusion models can memorize the training samples~\cite{somepalli2023diffusion,somepalli2023understanding}. Through empirical study and analysis, \citeauthor{somepalli2023understanding}~\cite{somepalli2023understanding} concluded that the diffusion models are more prone to memorize and replicate training samples during inference comparing to conventional generative adversarial networks.

Researchers leveraged membership inference attack (MIA)~\cite{duan2023diffusion,mia} to address the question of identifying which images in a collection containing private images were part of the training data. Additionally, \citeauthor{nicolas2023extracting}~\cite{nicolas2023extracting} assumed the collection didn't exist, knowing only prompts from the training data. They used these prompts to generate numerous images and successfully reconstructed some training samples using MIA\@. Although these assumptions are unrealistic for practical cases, they show that diffusion models are more vulnerable due to memorizing training samples.

\subsection{Data Reconstruction from Model Weights}
Multiple threats have been proposed specifically to reconstruct private input data from a given weights of a model. Apart from membership inference attacks which assume such a collection of images or prompts is known, there are two kinds of attacks without such an assumption. Given the weights of a trained model and a pre-trained model, the attacker first randomly initialized an input called dummy input and then conducted the exact same training process to update the pre-trained model. The attacker then compares the difference between the weights of the updated model and the given trained model and uses such a difference to update the dummy input through certain methods such as gradient descents. By minimizing such a difference, an attacker can reconstruct the private data which is represented by the dummy input.

Deep Leakage from Gradient~\cite{zhu2019deep} is the first paper to propose such a method. However, they can only apply to the case that the gradient of each step is shared during training.\ \citeauthor{geiping2020inverting}~\cite{geiping2020inverting} extended such a method to the case that the model weights are shared after training on the local.\ \citeauthor{buzaglo2024deconstructing}~\cite{buzaglo2024deconstructing} further extended such a method to more general cases where the optimizer for gradient descent is not limited to stochastic gradient descent optimizer. However, in our paper, we find such a method does not work over diffusion models and we further propose a new kind of attack.

\citeauthor{balle2022reconstructing}~\cite{balle2022reconstructing} studied another approach of reconstructing training samples with the attacker knowing the model weights. On the assumption that an adversary knows all the training data points except one, they built and trained reconstructor networks to reconstruct the missing data point. However, two factors hinder such a method from answering our research questions. First, they assumed a powerful and albeit unrealistic adversary~\cite{balle2022reconstructing}. Second, their method is designed for convex and classification models such as a three-layer convolution neural network having 4096 parameters.

\subsection{Defense Against Data Reconstruction}
To defend against data reconstruction from model weights, a typical method is adding random noise. The most common approach is differential privacy. DPDM~\cite{dockhorn2023differentially} used local differential privacy gradient descent in diffusion model training. DPGM~\cite{jiang2024functional} applied Renyi differential privacy for generative modeling, providing a better trade-off between privacy and image generation. We will study whether existing defense methods are effective against our attack.

\section{Motivation and Threat Model}
\subsection{Memorization During Fine-Tuning}\label{sec:memory}
It has been shown that diffusion models can memorize training samples. We would like to have a second thought on why we can reconstruct private images for fine-tuning with the given weights of a diffusion model. We notate the weights of the pre-trained model as $\theta$ and the weights of the fine-tuned model as $\theta'$. So, the model update is $\Delta \theta$. We notate the private samples for fine-tuning as $\mathcal{X}$. As we can use any letters as trigger words for a specific instance for fine-tuning, the instance is irrelevant to the trigger words. If we use different trigger words, we will have different $\Delta \theta$. Different fine-tuned models will take in different trigger words but they will output the images of the same identity. We can write this intuitive observation in the form of 
\begin{equation}
 \Pr(\Delta \theta |\mathcal{X},\theta)\Pr(\mathcal{X},\theta)=\Pr(\mathcal{X}|\Delta \theta,\theta)\Pr(\Delta \theta,\theta)
\end{equation}
The fine-tuning phase is a phase of memorizing private images so that we can generate images containing the special identity. As a result, it is intuitive that the information about private images are memorized in $\Delta \theta$. We will verify this argument empirically by showing the effectiveness of our proposed attack.
\subsection{Privacy Risks}
Previous studies have shown that diffusion models are vulnerable to attacks and can easily memorize training samples. However, privacy risks during the fine-tuning phase have not been thoroughly explored. Images used in fine-tuning can be more privacy-sensitive than those in the larger pre-training dataset, as they may include human faces, personal belongings, etc. Therefore, it is crucial to emphasize the potential privacy risks during fine-tuning.

\subsubsection{Definition of Privacy Leakage.} The previous literature defined privacy leakage as that the reconstructed images are exactly the same as training samples, for example having low mean square errors. We argue that as long as objects remain the same, privacy is compromised. For example, the faces of the same person are in reconstructed images and training samples. We argue that privacy leakage should not be limited to low mean square error between the two images. It should include a broader case that the same identity is revealed in the reconstructed and private images. For example, the reconstructed images have the faces of a person but not in the same position as in the private images.

\subsubsection{Assumptions in the Previous Literature.} In previous literature, they succeeded in reconstructing training samples. However, their assumptions are not valid enough to make such attacks have a chance to happen in the real world as their primary purpose is not to design the most practical privacy attack.\ \citeauthor{nicolas2023extracting}~\cite{nicolas2023extracting} assumed that the prompts in training data is already known to the adversary part. However, there is little chance that the users will use exactly the same prompt in the training for the inference nor will they share the training prompts with the others. During the fine-tuning phase, revealing the fine-tuning prompt is the same as exposing the private images as the prompt directly contains the trigger words. Regarding MIA~\cite{duan2023diffusion}, there is also little chance that an adversary is able to get a collection of images containing the same images in private dataset. It is said in the paper that the scope of MIA is limited to the real world.\ \citeauthor{balle2022reconstructing}~\cite{balle2022reconstructing} assumed an adversary knows all training data points except one. As stated in the literature, this assumption represents a powerful yet unrealistic adversary. Therefore, when fine-tuning a diffusion model, it's essential to establish more realistic assumptions regarding what information an adversary may possess.

\subsubsection{Our Assumptions.} Since leaking trigger words will directly reveal identities in private images, we should avoid sharing trigger words with untrusted devices. Unlike previous assumptions, we assume both prompts containing trigger words and private images used for fine-tuning are kept secret. However, the fine-tuned model weights are shared among devices.

One use case is that during the training phase of federated learning, each client downloads the pre-trained global model from the cloud and then fine-tunes the model with their local private data individually. A server then collects the weights of all these fine-tuned models and aggregates them into a global model so that the global model has superior performance than each locally fine-tuned model. In the mean time, each client does not need to share private images nor trigger words so that privacy is kept.

Another use case is fine-tuning a pre-trained diffusion model with private images. Sharing the fine-tuned model to a public or cloud hub is easy, for example, using the Hugging Face API with one line of code. Afterward, friends can use trigger words to compose prompts and request the cloud to generate images. However, the cloud cannot infer private images as it doesn't know the trigger words or which prompts contain them. Previous membership inference attacks won't help since the same prompts are unlikely to be used during fine-tuning. Thus, users and friends can do inference without revealing private images.

Another difference in our assumptions is that we assume the model structure is open-source. Unlike DALL-E-2 where only an application interface is provided, since the users are going to fine-tune the model with their private images, the model structure is transparent to them so that they can complete the fine-tuning.

As a result, \emph{our assumption} is that first of all, the pre-trained weights, model structures and algorithms of fine-tuning are open-source on the Internet. For the one who wants to reconstruct private images, it only has access to fine-tuned model weights of the same model. It does not have access to prompts including trigger words during the fine-tuning process and private images.

\subsection{Threat Model}
Our threat model considers an adversary reconstructing the same objects in the private images used for fine-tuning. Take \cref{fig:PrivateImagesElon} as an example, the adversary is not expected to reconstruct these six images. But if we are able to recognize that the reconstructed image contains the face of Elon Musk, the adversary achieves its goal.

The most common case of such an adversary is an honest-but-curious server where the server will complete the designated tasks such as aggregation in federated learning~\cite{fedavg}. But it will also try to reconstruct private data in the background process. This case is hard to detect. A terminology to describe this kind of threat model is \emph{informed adversaries}~\cite{balle2022reconstructing}. Data reconstruction attacks is the most serious attack in model inversion attacks~\cite{song2023real}. 

In our assumption, we do not consider models encrypted through trusted execution environments, multi-party computation, or homomorphic encryption. Our focus is on data privacy rather than data encryption. The overhead is not on the same scale as with plain data. For instance, the forward time on diffusion models with homomorphic encryption~\cite{chen2024privacypreserving} is 79.19 days compared to 35 seconds with plain text using NVIDIA A100. Furthermore, to the best of our knowledge, there is no encryption method that can be directly applied to the training process of diffusion models. We also assume that the fine-tuning is conducted through the method of LoRA\@.

\subsection{Quantify the Threats}\label{sec:quantify}
Previous literature used mean square error to assess image similarity. However, we assume that sharing the same identity constitutes privacy leakage. To quantitatively measure an adversary's success, we evaluate whether the same identity is revealed in two aspects.

Mentioned in DreamBooth~\cite{dreambooth}, one important aspect to evaluate is subject fidelity: preserving subject details in generated images.
\begin{itemize}
    \item CLIP-I measures the average pairwise cosine similarity (0 to 1) between CLIP embeddings of two images. A higher value indicates more private image details are preserved in the reconstruction, making the inversion attack more effective.
    \item CLIP-T measures average cosine similarity between prompt and image CLIP embeddings (0 to 100). Higher values indicate a more effective attack, as similarity between reconstructed and private images increases.
\end{itemize}

Another aspect is straightforward: measuring the recovery of private identities. For human faces, we first use private images to train a face recognition model, InceptionResNet V1~\cite{schroff2015facenet}. After the inversion attack, we test if the reconstructed images can be recognized by this model. Higher recognition accuracy indicates a more effective attack. InceptionResNet V1 is pre-trained on VGGFace2~\cite{cao2018vggface2}.

For common objects, we use DINO similarity. DINO~\cite{caron2021emerging} is a large self-supervised model trained on numerous natural images, utilizing the pre-trained ViT-S/16 model. This metric measures cosine similarity between feature embeddings before the last layer. Higher similarity between private and reconstructed images suggests a greater chance of recognizing the same object, indicating a more effective attack.
\section{Reconstructing Private Images Only With Model Weights}

\begin{figure}[tb]
	\centering
	\includegraphics[width=\linewidth]{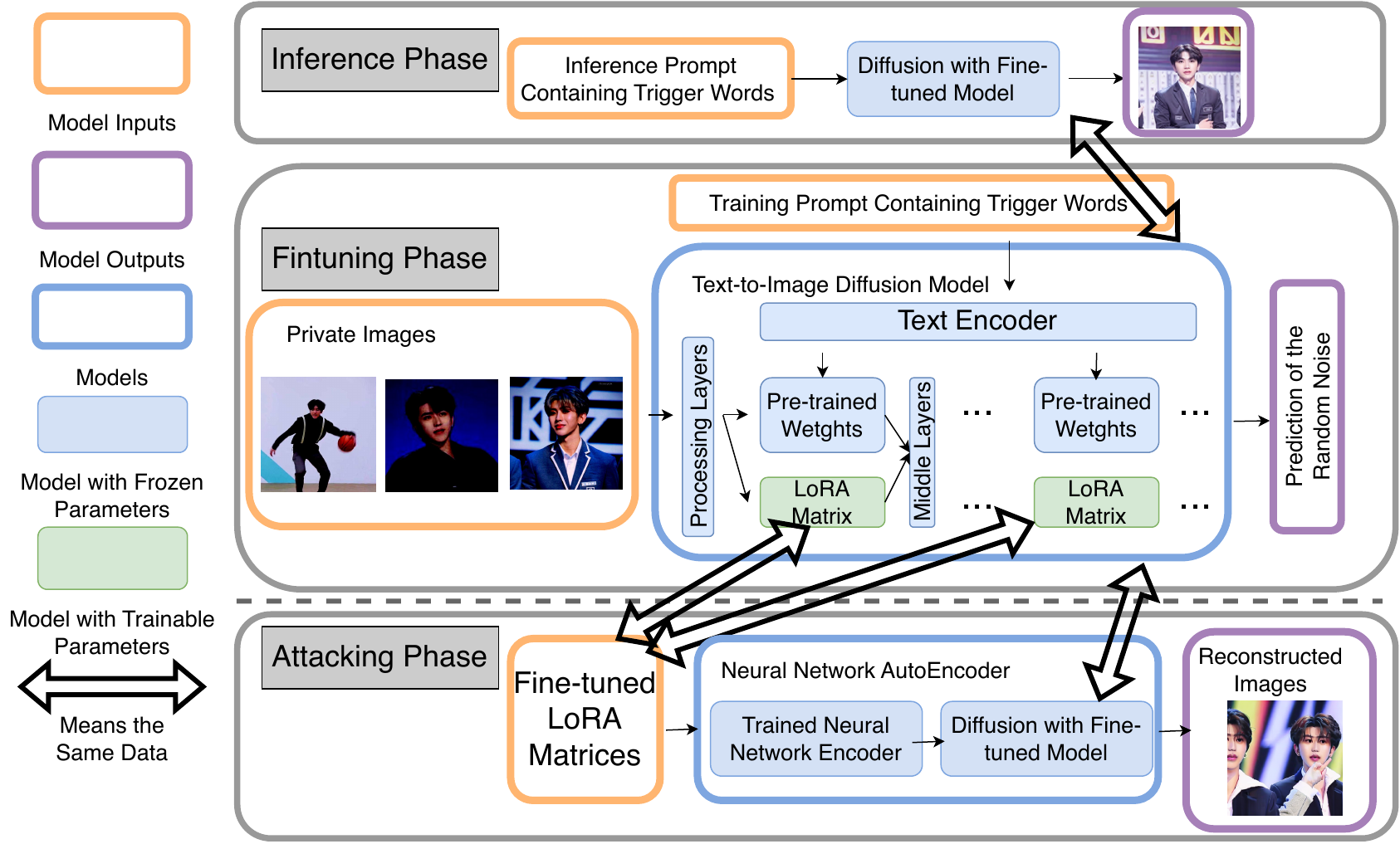}
	\caption{The overall framework of fine-tuning a diffusion model and our attack method}\label{fig:overall}
   \end{figure}

   For diffusion models, previous methods like matching gradients, model updates, or using reconstructor networks are ineffective for reconstructing private images. Consequently, we design an inversion network that takes $\Delta \theta$ as input and outputs the reconstructed image. Since $\Delta \theta$ from the LoRA method contains fewer fine-tuned parameters than full model fine-tuning, using it as input simplifies optimizing the inversion network.
   
   Illustrated in~\cref{fig:overall}, during the fine-tuning phase, we obtain LoRA matrices trained with private images. Simultaneously, an attacker will have an attacking model. Depicted in \cref{fig:overall}, the fine-tuning and private images reside on the user end, with only the LoRA matrices being shared with potential adversaries. The attacking model takes these trained LoRA matrices as input for a neural network encoder, which outputs the embedding of the fine-tuned model weights. This embedding replaces the text embedding output by a text encoder and is then fed into the fine-tuned diffusion model. Using the fine-tuned diffusion model, the attacker generates images with this embedding, considering the generated images as reconstructions of private images. These reconstructions are expected to have the same identity as the private images. In the following paragraphs, we will introduce how we design and build the inversion network and how an adversary trains it and uses it to launch the attack.
   
   \subsection{Build an Neural-Network-to-Image Autoencoder (Inversion Network)}
   \subsubsection{Neural Network Encoder.} Unlike conventional neural networks, which typically take images, videos, or texts as input, our neural network encoder accepts model weights as input and produces an embedding vector as output. The widely adopted LoRA method for fine-tuning allows us to use these model weights. For a given diffusion model, we only need to optimize the LoRA matrices, reducing the number of parameters to less than 1\% of the original trainable parameters. For instance, stable diffusion V-1.4~\cite{sd14} trainable parameters decrease from over 86 million to 1.6 million. We denote these matrices as $\Delta \theta$, which serve as inputs to the encoder.
   
\begin{figure}[tb]
	\centering
	\includegraphics[width=\linewidth]{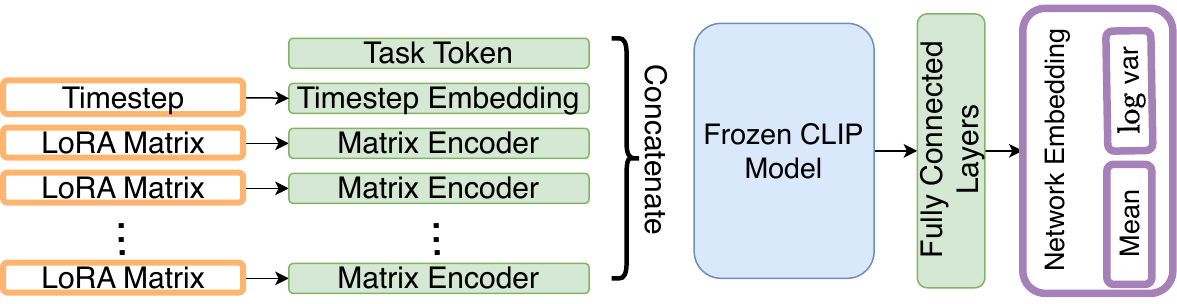}
	\caption{The structure of neural network encoder}\label{fig:encoder}
   \end{figure}

   We design the neural network encoder structure, shown in~\cref{fig:encoder}, based on two considerations. First, it should use as few parameters as possible while processing all LoRA matrices. Second, it should be easy to train and capture sufficient information. Therefore, we use several linear transformation layers and a large pre-trained model. The linear layers align the dimensions of input and output with the pre-trained model's requirements. The pre-trained model is fine-tuned to generate effective features. Further empirical study in \cref{sec:design} will explain our architectural choices.
   
\begin{figure}[tb]
	\centering
	\includegraphics[width=\linewidth]{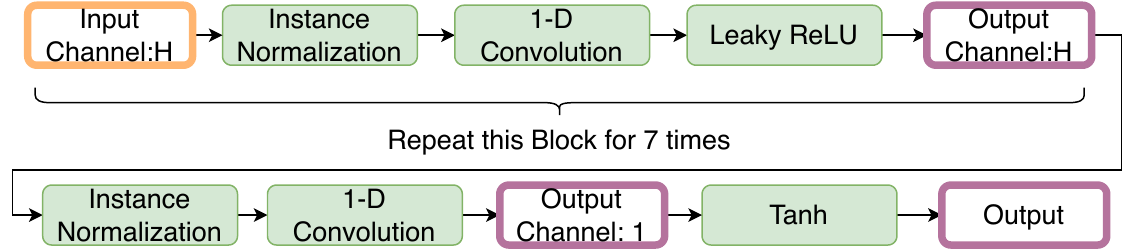}
	\caption{The structure of matrix encoder where corresponding LoRA matrix is in dimension of $W\times H$. The kernel size of convolution layer is 3.}\label{fig:matrixencoder}
   \end{figure}

   \textbf{Linear Transform Layers.} For each LoRA matrix, we use a matrix encoder to transform it into a vector. Since LoRA matrices vary in size, we need a separate encoder for each, with different input dimensions. The resulting vector is called a matrix embedding. The encoder consists of eight layers of 1-D convolutional layers with LeakyReLU activations and instance normalization. The width of the LoRA matrix determines the number of channels, with the last convolutional layer having a single channel. A linear layer then projects the output to a uniform dimension across all embeddings. The structure of the LoRA matrix encoder is shown in \cref{fig:matrixencoder}.
   
   In addition to the matrix embedding, we have a task token, a vector with the same dimension as the matrix embedding. We also have a timestep embedding, corresponding to the number of steps used to fine-tune a diffusion model. For example, with 1000 fine-tuning steps, the timestep embedding input is 1000, and the output is a vector. The timestep's usage will be introduced later, as it helps accelerate training of the inversion network. 
   
   \textbf{Large Pre-trained Model} We concatenate all these together and input them into a frozen CLIP model, as it has been pointed out that CLIP is a widely used model for encoding~\cite{clip}. Adopting the pre-trained CLIP model can help us better embed the network parameters into the domain of text embeddings. To align with the original embedding which are embedding of the texts, we use the text encoder of the CLIP model. The output serves as the embedding of updates of a neural network. The position embedding layer of the CLIP model needs to be re-trained, as the number of LoRA matrices is different from the original sequence length of the CLIP model. 
   
   \subsubsection{Complete Structure (Autoencoder):} Utilizing the neural network encoder, we construct an autoencoder. The neural network encoder, as depicted in \cref{fig:encoder}, incorporates two fully connected layers in the end, tasked with generating the mean and variance of the network embedding. Subsequently, we use these parameters to sample a network embedding from a normal distribution. For the diffusion model component, we directly utilize the fine-tuned diffusion model. The neural network embedding then replaces the text embedding in the original text-to-image model, while the remaining components remain unchanged.
   
   \subsection{Train the Inversion Network}
   Before the attackers initiate the attack with given model weights, they must first train a network encoder on a public dataset. This training occurs offline by the adversary before fine-tuning the diffusion models begins. The public dataset may not contain images of the exact private identities but should include closely related subjects. For example, if the private images are human faces, the encoder must be trained on a dataset with human faces or similar subjects. In practice, adversaries can train multiple inversion networks on various public datasets, such as those with nature images or human faces, and try each sequentially. Thus, training the encoder on a similar public dataset is practical.
   
   \begin{algorithm}[tb]
	\caption{The trivial method of training the network encoder}\label{alg:simpletrain}
	\begin{algorithmic}[1]
	\STATE{{\bfseries Input:} The pre-trained diffusion model $\theta_0$ and a public dataset.}\;
	\STATE{{\bfseries Output:} The trained network encoder with $M$ iterations.}\;
	\FOR{iterations $\leftarrow$ 1 to $M$}
	\STATE{Pick a batch of images from the public dataset: $\mathcal{X}$.}\;
	\FOR{fine-tuning rounds $r\leftarrow 1$ to $s$}
	\STATE{Generate Gaussian noise and add it to $\mathcal{X}$.}\;
	\STATE{Fine-tune the diffusion model one step from $\theta_{r-1}$ to $\theta_r$.}\;
	\ENDFOR{}
	\STATE{Calculate model updates: $\Delta \theta\leftarrow\theta_s-\theta_0$.}\;
	\STATE{Input $\Delta \theta$ into the network encoder and output the network embedding.}\;
	\STATE{Generate Gaussian noise and add it to $\mathcal{X}$.}\;
	\STATE{Input noisy images and network embedding into the fine-tuned diffusion model $\theta_s$.}\;
	\STATE{$\theta_s$ will output the prediction of Gaussian noise.}\;
	\STATE{Freeze $\theta_s$ and update the network encoder based on the difference between the ground truth and the prediction of the Gaussian noise.}\;
	\ENDFOR{}
	\end{algorithmic}
   \end{algorithm}
   
   The basic training paradigm, though complex, follows these steps, as shown in \cref{alg:simpletrain}. We assume $M$ training iterations. In each iteration, we reset the weights of a diffusion model to their pre-trained state. We then select a batch of training samples with the same identity and fine-tune the model for $s$ steps. We obtain the LoRA matrices, denoted by $\Delta \theta$, after fine-tuning. This outlines how we acquire one data point for the neural network encoder's training sample, as depicted from line 4 to line 9 in \cref{alg:simpletrain}.
   
   Next, we update the network encoder in each iteration (lines 10 to 14 in \cref{alg:simpletrain}). The input and output configurations are similar to the training process for fine-tuning a diffusion model. However, the input differs: while training a diffusion model involves a noisy image and a text embedding, here it involves a noisy image and a network embedding. The fine-tuned weights for the diffusion model remain frozen. We select the same batch of images (line 4), sample Gaussian noise to create noisy images, and input these into the fine-tuned diffusion model. The text embedding is replaced by the network embedding, which the diffusion model uses to predict the Gaussian noise. We then perform back-propagation to update the network encoder's parameters.
   
   As we can see, such a process is quite inefficient because for each step of updating the network encoder, we need $s$ steps of fine-tuning the diffusion model first. To carry out $\mathcal{O}(M)$ steps of updating the encoder, we need extra $\mathcal{O}(Ms)$ steps of gradient back-propagation.
   
   To expedite the training of the encoder, we introduce a timestep embedding into the composition of the network embedding. The enhanced algorithm is presented in \cref{alg:efficient}. We iterate through the $M$ steps of training. During each iteration of updating the encoder, we fine-tune a diffusion model for $s'$ steps instead of $s$ steps, where $s'$ is uniformly sampled from $[1,s]$. At each step of fine-tuning the diffusion model, we extract the model updates. These updates, along with the timestep (which represents the total steps used for fine-tuning the diffusion model up to that point), are inputted into the encoder. For instance, if the current fine-tuning round is $r$, the timestep is also $r$. The encoder then generates the embedding, and we update the encoder in the same manner (lines 13 to 16). This approach allows us to update the encoder for every step of fine-tuning the diffusion model. To perform $\mathcal{O}(M)$ steps of updating the encoder, we require an additional $\mathcal{O}(M)$ gradient back-propagation, significantly enhancing efficiency.
   
   \begin{algorithm}[tb]
	\caption{The efficient method of training the network encoder}\label{alg:efficient}
	\begin{algorithmic}[1]
	\STATE{{\bfseries Input:} The pre-trained diffusion model $\theta_0$ and a public dataset.}\;
	\STATE{{\bfseries Output:} The trained network encoder with $M$ iterations.}\;
	\STATE{\textit{iteration} $\leftarrow$ 0.}\;
	\WHILE{\textit{iteration} $<M$}
	\STATE{Pick a batch of images from the public dataset: $\mathcal{X}$.}\;
	\STATE{Uniformly sample $s'$ from $[1,s]$.}\;
	\FOR{fine-tuning rounds $r\leftarrow 1$ to $s'$}
	\STATE{\textit{iteration} $\leftarrow$ \textit{iteration} $+1$.}\;
	\STATE{Generate Gaussian noise and add it to $\mathcal{X}$.}\;
	\STATE{Fine-tune the diffusion model one step from $\theta_{r-1}$ to $\theta_r$.}\;
	\STATE{Calculate model updates: $\Delta \theta\leftarrow\theta_r-\theta_0$.}\;
	\STATE{Input $\Delta \theta$ and timestep $r$ into the network encoder and output the network embedding.}\;
	\STATE{Generate Gaussian noise and add it to $\mathcal{X}$.}\;
	\STATE{Input noisy images and network embedding into the fine-tuned diffusion model $\theta_s$.}\;
	\STATE{$\theta_s$ will output the prediction of Gaussian noise.}\;
	\STATE{Freeze $\theta_r$ and update the network encoder based on the difference between the ground truth and the prediction of the Gaussian noise.}\;
	\ENDFOR{}
	\ENDWHILE{}
	\end{algorithmic}
	\end{algorithm}

   \subsection{Launch the Attack}
   To initiate the attack, we combine the offline-trained neural network encoder and the fine-tuned diffusion model into a network autoencoder. We generate the network embedding and use it in place of the presumed text embedding. The timestep input into the encoder represents the total steps used for fine-tuning the diffusion models, denoted as $s$. We then generate an image using the diffusion model, following the attack phase depicted in \cref{fig:overall}. The image generation process is similar to that of a conditional diffusion model, except the text embedding is replaced by the network embedding. The resulting image reconstructs private images. While controlling specific content may be challenging, the identity will remain consistent with the private images.
   
\section{Evaluation of the Attack}
We will first evaluate the effectiveness of the proposed attack and compare it with other baselines. In this section, we will not consider particular defense methods but assume that the model weights are shared in forms of plain text.

\subsection{Experiment Settings}
\subsubsection{Models and Datasets.} We use stable diffusion model V-1.4, with pre-trained weights downloaded from the Internet~\cite{sd14}. The text encoder is a CLIP ViT-L/14 model~\cite{clipdownload}. The frozen CLIP model used in the neural network encoder is also a CLIP ViT-L/14~\cite{clipdownload} model. The input resolution is 512$\times$512, and all images are scaled to this resolution. The CLIP text encoder is 241.03 MB, and our entire encoder in the inversion network is 444.51 MB\@. Despite the large dimensions of LoRA matrices, our design ensures a light-weighted encoder, reducing overfitting risks as discussed in \cref{sec:fittext}.

As the most privacy-sensitive data consist of human faces, we selected the CelebAHQ dataset~\cite{CelebAMask-HQ}, which contains 30,000 high-resolution face images and 6,217 different identities, as the most representative dataset. We chose 5,000 identities as training samples for the encoder and the remaining images as the validation set. No identity appears in both the training and validation sets, ensuring features in the validation set are never encountered during training.

In addition, to demonstrate that our method also functions effectively with datasets beyond human faces, we have selected ImageNet~\cite{deng2009imagenet} as the public dataset and the dataset provided by DreamBooth~\cite{dreambooth} as the validation set for private images. There is no overlap in identities between these two datasets.

\begin{table}[tb]
	\caption{Effectiveness of various attack methods when using images from the CelebAHQ and DreamBooth datasets to fine-tune a stable diffusion V-1.4 model.}\label{tab:attackcelebhq}
	\begin{center}
		\begin{threeparttable}
		\begin{tabular}{cccc}
			\toprule
			\multicolumn{4}{c}{CelebAHQ}\\
			\midrule
			&CLIP-T&CLIP-I&Rec. Acc.\\
			\midrule
			DreamBooth&25.54&0.88&0.99\\
			\midrule
			\multicolumn{4}{c}{Attacks}\\
			\midrule
			Ours &25.41&0.87&0.99\\
			\citeauthor{geiping2020inverting}&23.20&0.75&0.010\\
			\citeauthor{buzaglo2024deconstructing}&22.13&0.68&0.013\\
			\midrule
			\multicolumn{4}{c}{DreamBooth}\\
			\midrule
			&CLIP-T&CLIP-I&DINO-Similarity\\
			\midrule
			DreamBooth&25.47&0.95&0.88\\
			\midrule
			\multicolumn{4}{c}{Attacks}\\
			\midrule
			Ours &24.57&0.93&0.86\\
			\citeauthor{geiping2020inverting}&24.86&0.78&0.68\\
			\citeauthor{buzaglo2024deconstructing}&20.67&0.80&0.77\\
			\bottomrule
		\end{tabular}
\end{threeparttable}
\end{center}
\end{table}

\subsubsection{Baselines.} There are two baselines regarding reconstructing private images from model weights. The first one is proposed by \citeauthor{geiping2020inverting}\cite{geiping2020inverting}. We use this method to find a dummy input that minimizes the difference between given model updates and fine-tuned model updates optimized from the dummy input. The second method is proposed by \citeauthor{buzaglo2024deconstructing}\cite{buzaglo2024deconstructing}. We choose the method of data reconstruction with general loss functions because our loss function for fine-tuning is MSE loss. For these two baselines, as the embedding layer of embedding word tokens into features cannot propagate the gradients, we actually reconstruct the embedding of tokens rather than the original input tokens.

\subsubsection{Fine-Tuning and Training Settings.}
As LoRA is a common method for fine-tuning large generative models, we use it as our default. For fine-tuning a diffusion model with LoRA, we set the steps $s$ to 1000. According to DreamBooth, we generate 200 class images as negative samples with prompts like `A [C]', where C is the class name (e.g., human face or dataset labels). Trigger words are random strings of 4 to 11 letters from the lowercase alphabet. During fine-tuning, the prompt for private images is `A [V] [C]', and for testing, it is `An image of [V] [C] with high quality', with a negative prompt of `Low quality, distorted, weird images'. The pre-trained model weights remain fixed, while LoRA matrices are randomly initialized for each fine-tuning session.

The remaining settings are the same as those in the DreamBooth paper~\cite{dreambooth}. The text encoder is also fine-tuned. For the diffusion model, the rank is 16 and $\alpha$ is 27. For the text encoder, the rank is 16 and $\alpha$ is 17. We use the AdamW optimizer with a learning rate of $1\times 10^{-4}$ and gradient clip of 1.0. The weight decay is 0.01, and $\epsilon$ for the optimizer is $1\times 10^{-8}$.

To train the neural network encoder, we have $M = 100000$ training steps. We use the AdamW optimizer with a learning rate of 0.005. The weight decay is 0.0005, and the $\epsilon$ of the optimizer is $1\times 10^{-8}$. According to the stable diffusion model V-1.4, the dimension of the embedding is 768.

For the settings of baselines, we initialize the dummy input following a standard normal distribution ($\mathcal{N} (0,1)$). The optimizer used for updating the dummy input is Adam with a default learning rate of 0.01~\cite{geiping2020inverting}, with AMSGrad set to true. For fair comparison, we also perform $M=100000$ steps of attacking to ensure that the loss between given model updates and model updates trained from the dummy input converges.

\begin{table}[tb]
	\caption{Effectiveness of baseline attack methods when we are using images CelebAHQ dataset to fine-tune a stable diffusion V-1.4 model.}\label{tab:attacklr}
	\begin{center}
	\begin{threeparttable}
		\begin{tabular}{ccccc}
			\toprule
			$lr$&Attacks&CLIP-T&CLIP-I&Rec.~Acc.\\
			\midrule
			\multirow{2}{*}{0.1}&\citeauthor{geiping2020inverting}&22.32&0.76&0.015\\
			&\citeauthor{buzaglo2024deconstructing}&23.01&0.73&0.013\\
			\midrule
			\multirow{2}{*}{0.001}&\citeauthor{geiping2020inverting}&20.06&0.72&0.024\\
			&\citeauthor{buzaglo2024deconstructing}&21.72&0.79&0.031\\
			\midrule
			\multirow{2}{*}{0.0001}&\citeauthor{geiping2020inverting}&22.32&0.75&0.014\\
			&\citeauthor{buzaglo2024deconstructing}&20.93&0.75&0.031\\
			\bottomrule
		\end{tabular}
	\end{threeparttable}
	\end{center}
\end{table}

\subsubsection{Evaluation Metrics.} 
For numerical metrics, according to the analysis in \cref{sec:quantify}, we use those metrics to evaluate the effectiveness of the attack. Regarding the face recognition, if the threashold exceeds 0.9, it indicates that the two images represent the same person~\cite{schroff2015facenet}. `Rec. Acc.~' is the acronym for recognition accuracy. In presenting our results, we showcase the similarities between the reconstructed images and private images. Typically, when fine-tuning a diffusion model, we utilize 3 to 6 private images. Each attack generates one reconstructed image. 

In addition to numerical results, we provide example figures of private and reconstructed images within a practical case. We directly collect several images of Elon Musk's face from the Internet and consider them as private images for fine-tuning. Despite their differences, these images depict the same person. Notably, Elon Musk's faces are not included in the CelebAHQ dataset. Such visualizations better help understand the effectiveness of the attacks.

Furthermore, we measure the similarities between private images and images generated by the fine-tuned diffusion model to establish a benchmark. High similarities in this comparison indicate effective fine-tuning of the diffusion model. We use the acronym `DreamBooth' to denote these results, as they stem from images generated directly from models fine-tuned with DreamBooth.

\subsection{Effectiveness of Attacks}
\subsubsection{Baseline Attacks.} We are going to show the effectiveness of different attack methods without considering the defense methods. The two baseline attack methods on the other hand, are not so effective in reconstructing private data. Apart from trying the default values of learning rates, we also test other learning rate values ranging from $0.1$ to $0.0001$. The results are shown as in \cref{tab:attacklr}. We can see that the other two attacks are still ineffective. The reason why previous attack methods do not work is that the training process of fine-tuning a diffusion model is much more complicated and a lot of uncertain elements are involved such as class images. The gradients used to match model updates cannot be correctly calculated and back-propagated toward dummy inputs.

\subsubsection{Our Attack.} \Cref{tab:attackcelebhq} presents the performance of attacks on the CelebAHQ and DreamBooth datasets. The fine-tuned model effectively embeds specific identities into the images. Our attack method successfully reconstructs the same identities in private images with similarity scores close to those of DreamBooth. Notably, on the CelebAHQ dataset, the similarity between our reconstructed and private images exceeds 0.9, indicating successful reconstruction of the same identity. Therefore, our attack method proves effective in reconstructing private information.

To provide a straightforward assessment of our method's effectiveness, we also present some visualization results here.\ \cref{fig:PrivateImages} displays various types of private images. The goal for an attacker is to reconstruct either the face of Elon Musk or the yellow toy duck using only fine-tuned model weights.

\begin{figure}[tb]
	\centering
	\begin{subfigure}{0.24\linewidth}
		\includegraphics[width=\linewidth]{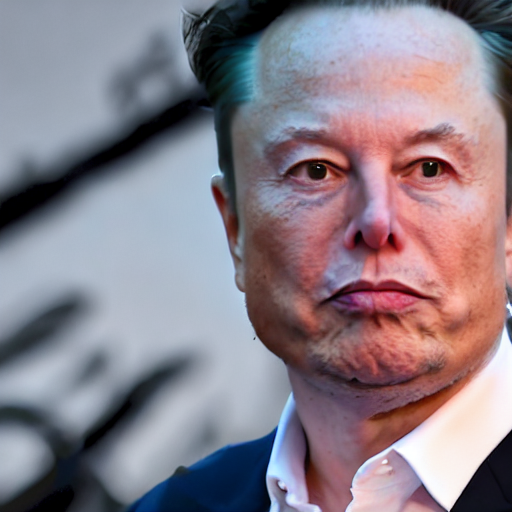}
		\caption{DreamBooth}
	\end{subfigure}
	\begin{subfigure}{0.24\linewidth}
		\includegraphics[width=\linewidth]{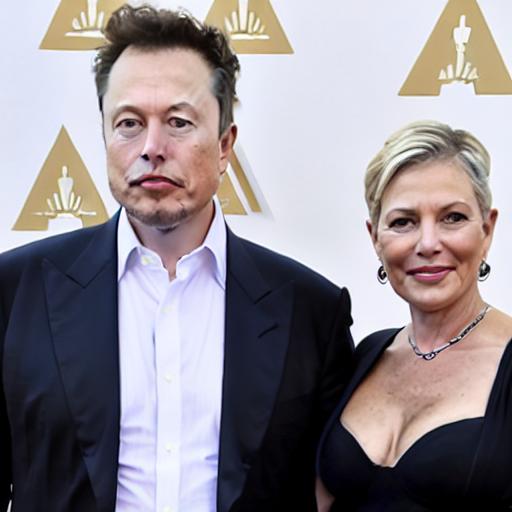}
		\caption{Ours}
	\end{subfigure}
	\begin{subfigure}{0.24\linewidth}
		\includegraphics[width=\linewidth]{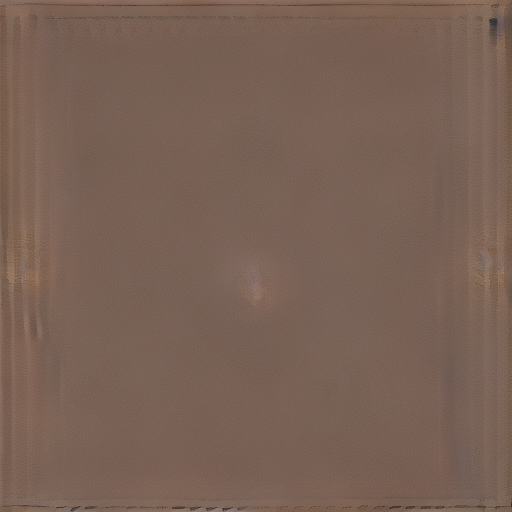}
		\caption{\citeauthor{geiping2020inverting}}
	\end{subfigure}
	\begin{subfigure}{0.24\linewidth}
		\includegraphics[width=\linewidth]{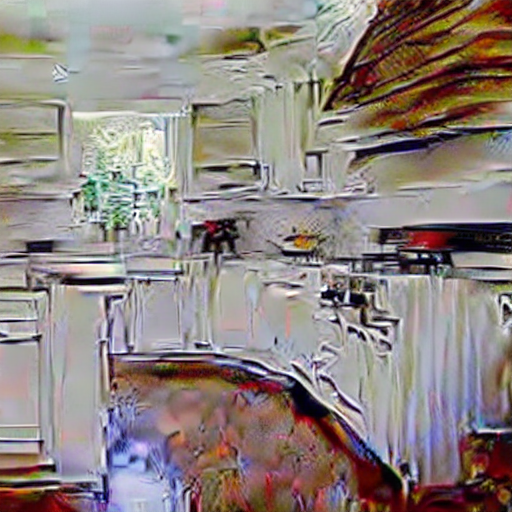}
		\caption{\citeauthor{buzaglo2024deconstructing}}
	\end{subfigure}
	\caption{The generated image by stable diffusion V-1.4 fine-tuned with images of Elon Musk and reconstruction by different attack methods.}\label{fig:attackMusk}
\end{figure}

In \cref{fig:attackMusk}, we observe that adversaries employing our method successfully reconstruct images containing Elon Musk with quality comparable to those generated directly by DreamBooth. Conversely, the other two baselines fail to generate the correct images and lack information about the private images.

\begin{figure}[tb]
	\centering
	\begin{subfigure}{0.24\linewidth}
		\includegraphics[width=\linewidth]{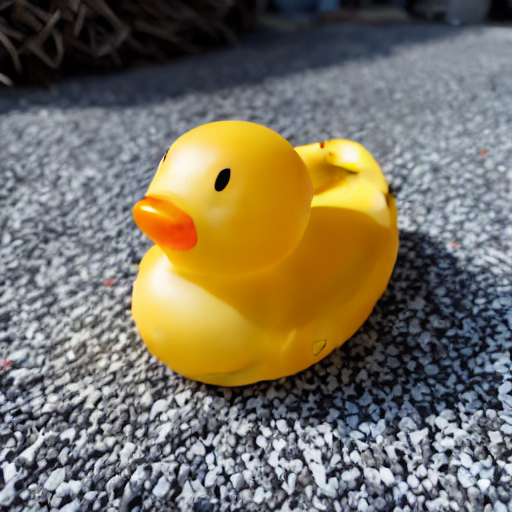}
		\caption{DreamBooth}
	\end{subfigure}
	\begin{subfigure}{0.24\linewidth}
		\includegraphics[width=\linewidth]{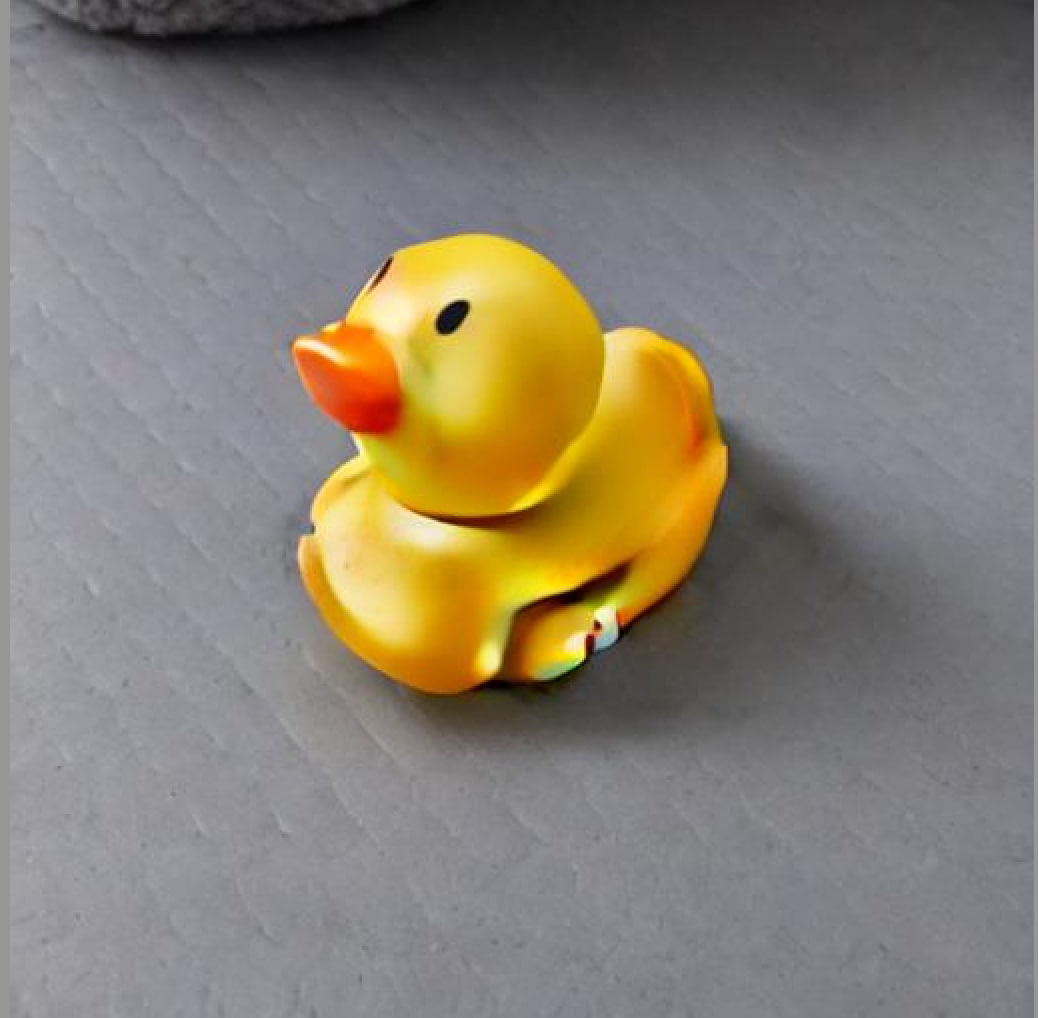}
		\caption{Ours}
	\end{subfigure}
	\begin{subfigure}{0.24\linewidth}
		\includegraphics[width=\linewidth]{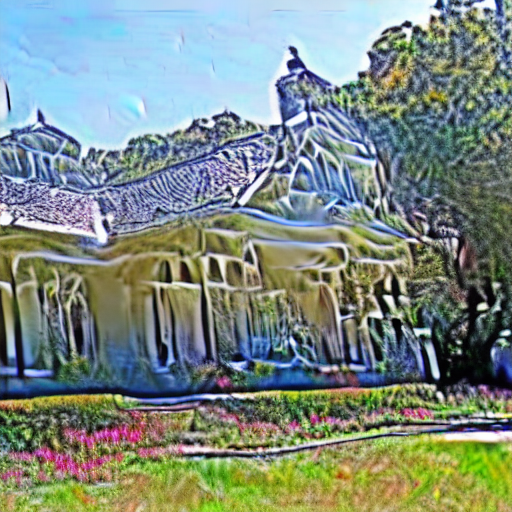}
		\caption{\citeauthor{geiping2020inverting}}
	\end{subfigure}
	\begin{subfigure}{0.24\linewidth}
		\includegraphics[width=\linewidth]{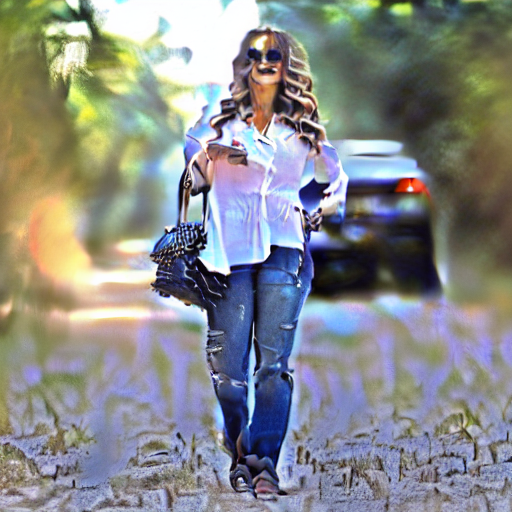}
		\caption{\citeauthor{buzaglo2024deconstructing}}
	\end{subfigure}
	\caption{The generated image by stable diffusion V-1.4 fine-tuned with images of toy duck and reconstruction by different attack methods.}\label{fig:attackDuck}
\end{figure}

From \cref{fig:attackDuck}, we can draw a similar conclusion. Although the reconstructed images by the attacker exhibit slight distortion, we successfully reveal the private identity. In contrast, images reconstructed by previous methods are completely different from the content in private images. In conclusion, we demonstrate that private images can be reconstructed solely using model weights.

\subsubsection{Effectiveness over Other User Model.} In the previous evaluation, we used stable diffusion V-1.4 as the user model. Here, we assess our attack's effectiveness on other user models. Since representative models like DALL-E-2 and Imagen~\cite{saharia2022photorealistic} are not open-sourced, we use stable diffusion V-2.1 as the alternative.

Stable diffusion V-2.1 differs from V-1.4 in several ways. V-2.1 uses OpenCLIP~\cite{openclip} as the text encoder, while V-1.4 uses the CLIP model, meaning the text embedding spaces are different. This difference affects our attack method, which substitutes text embeddings with network embeddings. Additionally, V-2.1 employs a training dataset with a less strict NSFW filter compared to the LAION-5B dataset used in V-1.4, leading to differences in the pre-training dataset. Moreover, V-2.1 is a more powerful model with an embedding dimension of 1024, necessitating a different inversion network. We aim to determine if these factors impact attack performance.

\begin{figure}[tb]
	\centering
	\begin{subfigure}{0.24\linewidth}
		\includegraphics[width=\linewidth]{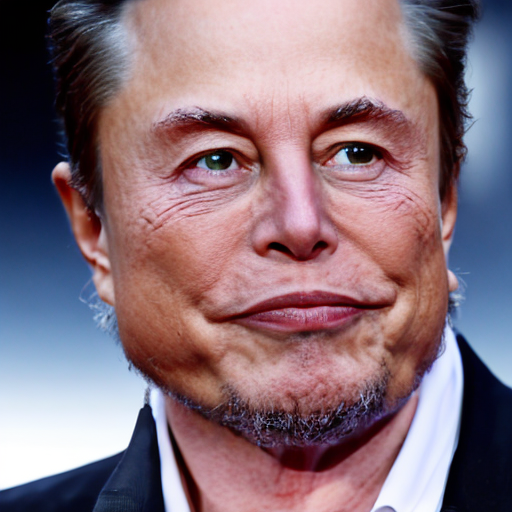}
		\caption{DreamBooth}
	\end{subfigure}
	\hfill
	\begin{subfigure}{0.24\linewidth}
		\includegraphics[width=\linewidth]{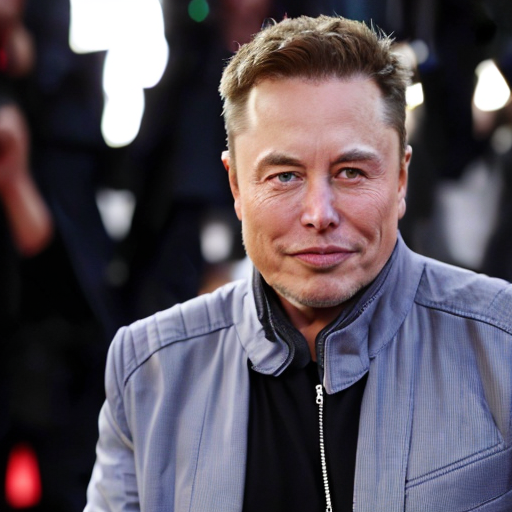}
		\caption{Ours}
	\end{subfigure}
	\hfill
	\begin{subfigure}{0.24\linewidth}
		\includegraphics[width=\linewidth]{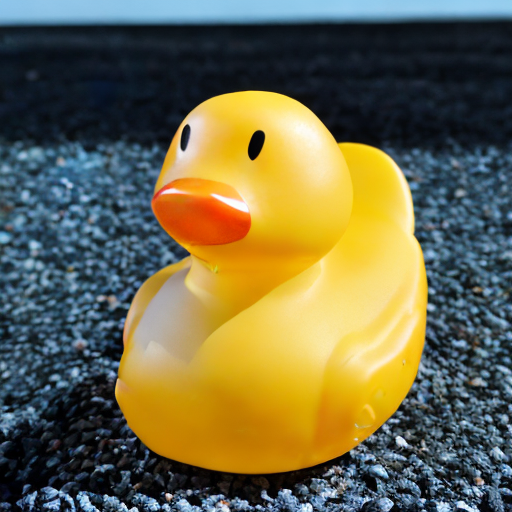}
		\caption{DreamBooth}
	\end{subfigure}
	\hfill
	\begin{subfigure}{0.24\linewidth}
		\includegraphics[width=\linewidth]{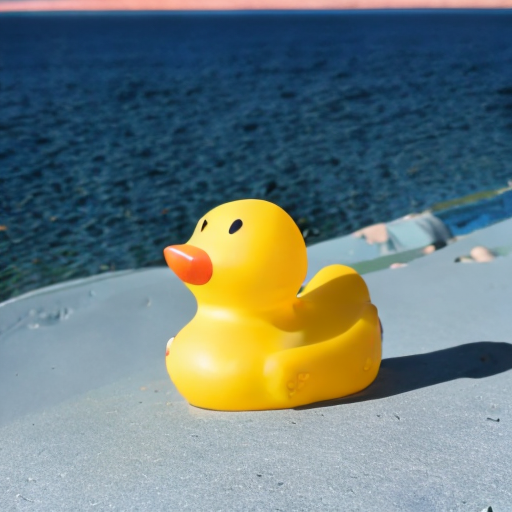}
		\caption{Ours}
	\end{subfigure}
	\caption{The generated image by stable diffusion V-2.1 and reconstruction by our attack.}\label{fig:attackSD2}
\end{figure}

From \cref{tab:attackcelebhqSD2}, we observe that even when we employ another diffusion model with different pre-training conditions, our attack can still reconstruct images with similarity scores close to those directly generated by the fine-tuned diffusion model. In \cref{fig:attackSD2}, we also successfully reconstruct images containing the expected identities. Therefore, we can conclude that the three factors we have listed do not affect the effectiveness of our attack. Our attack remains effective across different types of diffusion models.

\begin{table}[tb]
	\caption{The evaluation metrics to show the effectiveness of our attack when we use images in CelebAHQ and DreamBooth dataset to fine-tune a stable diffusion V-2.1 model.}\label{tab:attackcelebhqSD2}
	\begin{center}
		\begin{threeparttable}
		\begin{tabular}{cccc}
			\toprule
			\multicolumn{4}{c}{CelebAHQ}\\
			\midrule
			&CLIP-T&CLIP-I&Rec.~Acc.\\
			\midrule
			DreamBooth&24.52&0.90&0.98\\
			Ours &24.64&0.89&0.97\\
			\midrule
			\multicolumn{4}{c}{DreamBooth}\\
			\midrule
			&CLIP-T&CLIP-I&DINO-Similarity\\
			\midrule
			DreamBooth&22.23&0.90&0.60\\
			Ours &22.16&0.89&0.60\\
			\bottomrule
		\end{tabular}
	\end{threeparttable}
	\end{center}
\end{table} 

\subsection{Study on the Design of Network Encoder}\label{sec:design}
One core contribution of our work is that our design of the inversion network allows our attack to eliminate unrealistic assumptions and handle complex user models, such as diffusion models. This contrasts with previous literature, which often relies on reconstructing networks to reconstruct private data. In this section, we aim to address why such a design of a network encoder is necessary.

\subsubsection{Ablation Study on CLIP Model.}
The initial concept of building an inversion network is to adopt the approach used in constructing a reconstructor network~\cite{balle2022reconstructing}. However, there are significant challenges. Firstly, we cannot utilize the same reconstructor network, as its input size is only 4096. Flattening all the LoRA matrices used for fine-tuning stable diffusion model V-1.4 would result in over 1.6 million input units. Constructing or computing such a large fully-connected layer is impractical. Hence, the alternative approach is to use a matrix encoder for each LoRA matrix and encode each matrix into a vector. This enables us to encode matrices into an embedding with a similar format to a text embedding.

It seems that we can directly use such concatenated embeddings as a substitution for text embeddings. Therefore, the first question arises: why do we need a frozen CLIP model? The original CLIP model aimed to unify image and text modalities within a shared embedding space. Consequently, the primary rationale for utilizing a frozen CLIP model is to leverage the advantages of pre-training on large datasets to map the concatenated embeddings onto the embedding space of the original text embeddings.

\begin{figure}[tb]
	\centering
	\begin{subfigure}{0.24\linewidth}
		\includegraphics[width=\linewidth]{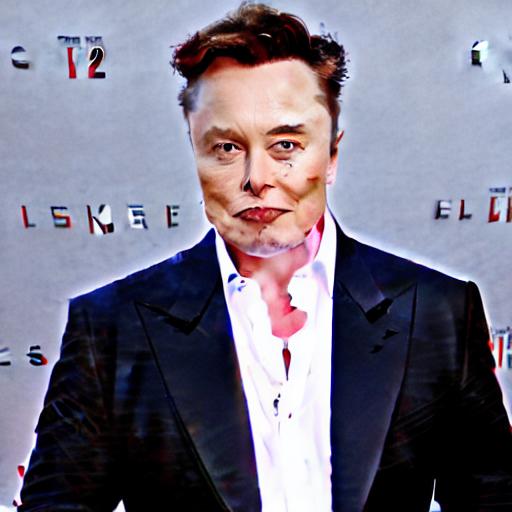}
		\caption{Elon Musk}
	\end{subfigure}
	\begin{subfigure}{0.24\linewidth}
		\includegraphics[width=\linewidth]{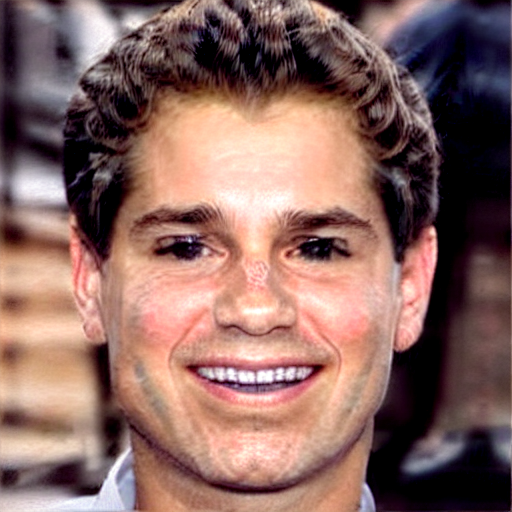}
		\caption{w/o CLIP}
	\end{subfigure}
	\begin{subfigure}{0.24\linewidth}
		\includegraphics[width=\linewidth]{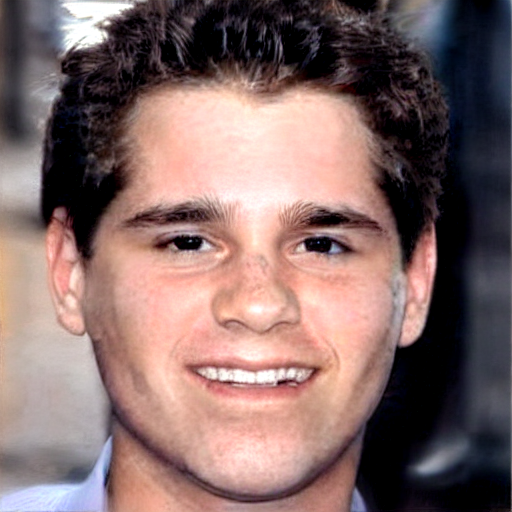}
		\caption{w/ CLIP}
	\end{subfigure}
	\begin{subfigure}{0.24\linewidth}
		\includegraphics[width=\linewidth]{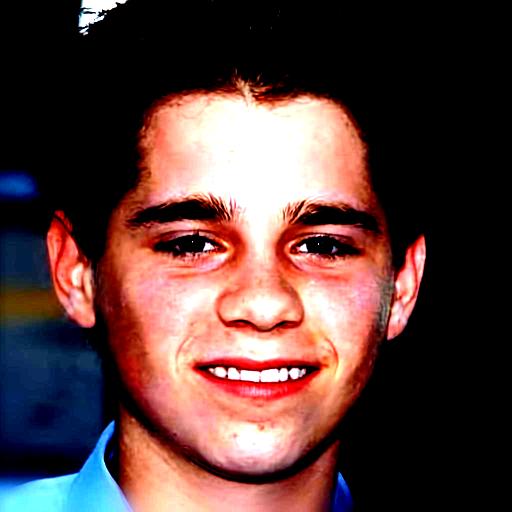}
		\caption{Private}
	\end{subfigure}
	\caption{The first image is the reconstruction result of Elon Musk by leveraging our attack but no having the frozen CLIP model. The right three images are the reconstruction by our attack without and with frozen CLIP model and an example of private image of a randomly picked identity of the validation set from CelebAHQ.}\label{fig:designNoCLIP}
\end{figure}

We do an experiment where we removed the frozen CLIP model and directly used the concatenated embeddings of the vectors encoded from the LoRA matrix to generate the network embeddings. Under such conditions, the CLIP-T score is 24.20, the CLIP-I score is 0.73, and the recognition accuracy is 0.65. From \cref{fig:designNoCLIP}, we can observe that the embeddings are not well-mapped to the embedding space matching the diffusion model best. The generated images deviate from the expected performance in correctly identifying identities. Some facial features are not accurately generated.

\subsubsection{Ablation Study on Matrix Encoder}
In \cref{fig:matrixencoder}, we use the 1-D convolution as the main component for processing the parameters of the LoRA matrices. Rather than the 1-D convolution layer, another widely-adopted structure is the linear layer, which is worked as a structure to encode the model parameters. To answer the question whether we can use the linear layer as the alternative as 1-D CNN, we replace the the convolution layers with the linear layers where the input dimension and output dimension is $W\times H$. $W\times H$ is the size of each LoRA matrix.

\begin{figure}[tb]
	\centering
	\begin{subfigure}{0.24\linewidth}
		\includegraphics[width=\linewidth]{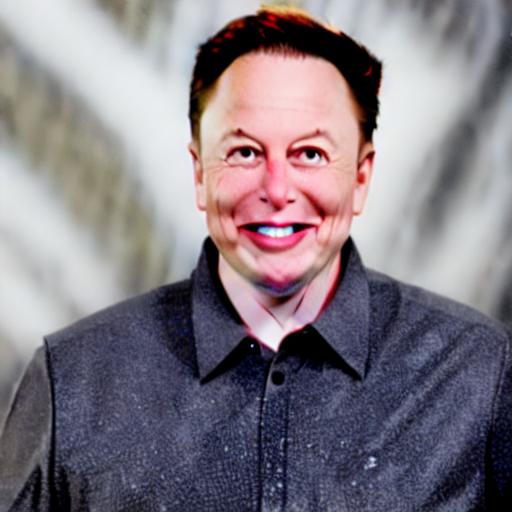}
		\caption{Elon Musk}
	\end{subfigure}
	\begin{subfigure}{0.24\linewidth}
		\includegraphics[width=\linewidth]{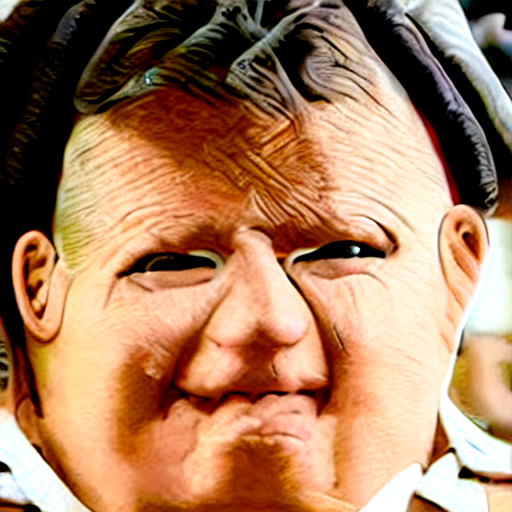}
		\caption{Linear}
	\end{subfigure}
	\begin{subfigure}{0.24\linewidth}
		\includegraphics[width=\linewidth]{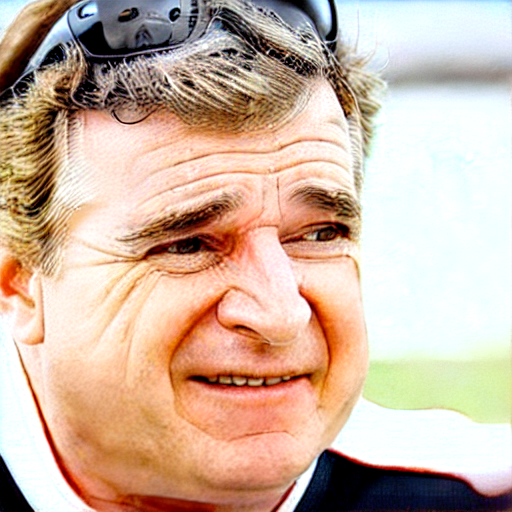}
		\caption{1-D CNN}
	\end{subfigure}
	\begin{subfigure}{0.24\linewidth}
		\includegraphics[width=\linewidth]{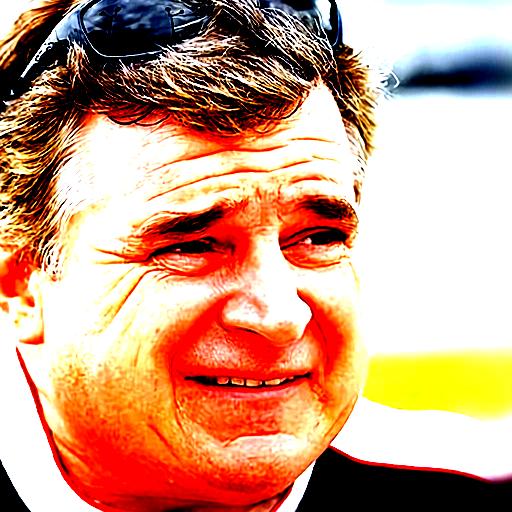}
		\caption{Private}
	\end{subfigure}
	\caption{The first image depicts the reconstruction result of Elon Musk achieved through our attack, utilizing the Linear layer as the primary component of the matrix encoder. The remaining three images showcase reconstructions obtained without and with the frozen CLIP model, along with an example of a private image featuring a randomly selected identity from the validation set of CelebAHQ.}\label{fig:designMatrixEncoder}
\end{figure}

The CLIP-T score is 22.13, the CLIP-I score is 0.74, and the recognition accuracy is 0.45. From \cref{fig:designMatrixEncoder} and the similarity scores, it's evident that performance with linear layers is insufficient. While the inversion network reconstructs some attributes of the expected identity, it fails with the private images. One reason is the difficulty in optimizing linear layers, and they also consume more memory. Thus, we conclude that using convolutional layers is better

\subsubsection{Ablation Study on the Number of LoRA Matrices Used}
According to the analysis in \cref{sec:memory}, we find that $\Delta \theta$ is determined by the images and prompts used to fine-tune the diffusion model. Given that each LoRA matrix participates in the fine-tuning process and its parameters are updated, it's evident that the LoRA matrix should contain information about the private images. Hence, the next question arises: why do we need to use all LoRA matrices to generate the network embedding?

\begin{table}[tb]
	\caption{The evaluation metrics to verify the effectiveness of baseline attack methods when we are using images CelebAHQ dataset to fine-tune a stable diffusion V-1.4 model.}\label{tab:designnumber}
	\begin{center}
		\begin{threeparttable}
		\begin{tabular}{cccc}
			\toprule
			\# LoRA matrices used&CLIP-T&CLIP-I&Rec.~Acc.\\
			\midrule
			1&23.35&0.73&0.011\\
			32&25.28&0.78&0.88\\
			64&25.24&0.85&0.91\\
			96&25.83&0.85&0.96\\
			\bottomrule
		\end{tabular}
	\end{threeparttable}
	\end{center}
\end{table}

\begin{figure}[tb]
	\centering
	\begin{subfigure}{0.24\linewidth}
		\includegraphics[width=\linewidth]{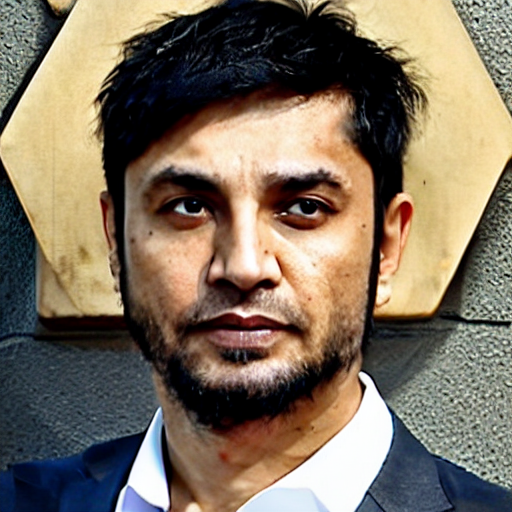}
		\caption{1}
	\end{subfigure}
	\begin{subfigure}{0.24\linewidth}
		\includegraphics[width=\linewidth]{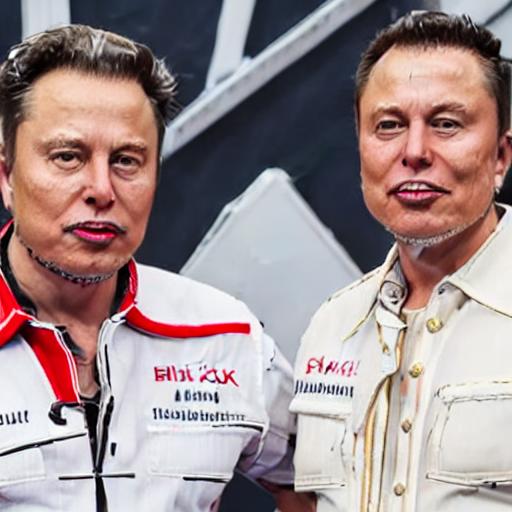}
		\caption{32}
	\end{subfigure}
	\begin{subfigure}{0.24\linewidth}
		\includegraphics[width=\linewidth]{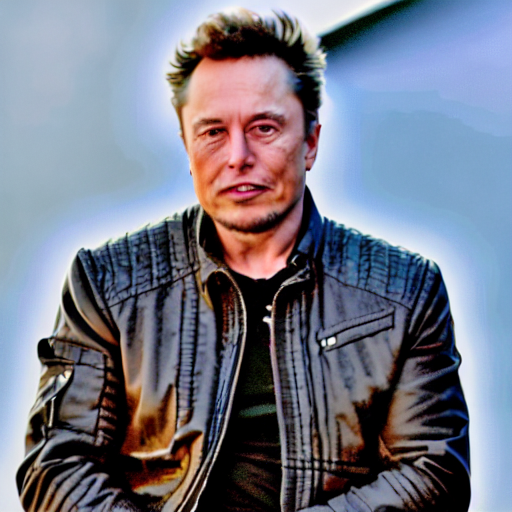}
		\caption{64}
	\end{subfigure}
	\begin{subfigure}{0.24\linewidth}
		\includegraphics[width=\linewidth]{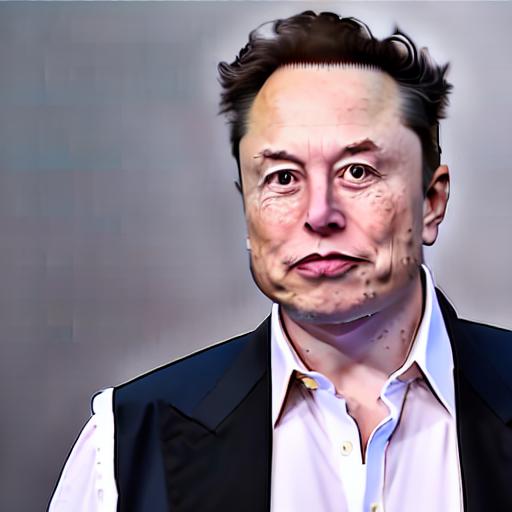}
		\caption{96}
	\end{subfigure}
	\caption{The reconstruction image of Elon Musk by using our attack method with different number of LoRA matrices}\label{fig:designNumberMatrix}
\end{figure}

For the stable diffusion model V-1.4, there are 128 matrices in total. In an experiment, we used only the first 1, 32, 64, and 96 LoRA matrices to generate the network embedding. \Cref{tab:designnumber} presents similarity scores measured using CelebAHQ with different numbers of LoRA matrices. During both the training and attack phases, we encode these numbers of LoRA matrices into the network embedding. \Cref{fig:designNumberMatrix} shows the reconstruction images of Elon Musk.

When only the first LoRA matrix is used, the generated result of the attacking reconstruction is completely random. However, when one fourth of all matrices are used, the reconstruction images already bear a resemblance to the original identity in the private images, albeit with some details lacking. Using 96 out of 128 matrices yields quite satisfactory results, capable of reconstructing the private identity. To achieve the best performance in attacking, we choose to use all LoRA matrices. Additionally, using all matrices improves the ease of use of our method, eliminating the need for others to determine how many parameters of the model weights are required for the attack. The standard practice is to use all matrices, eliminating the need for additional hyper-parameters.

\subsubsection{Ablation Study on Variational AutoEncoder}
In our network encoder, we use a linear layer to generate the mean value and another to generate the $\log$ of variance, which are then used to produce the embedding. This design is typical in a variational autoencoder. The key difference is that, instead of encoding an input as a single point, we encode it as a distribution over the latent space. This approach helps mitigate overfitting and allows the neural network encoder to represent the networks in a larger embedding space.

\begin{figure}[tb]
	\centering
	\begin{subfigure}{0.24\linewidth}
		\includegraphics[width=\linewidth]{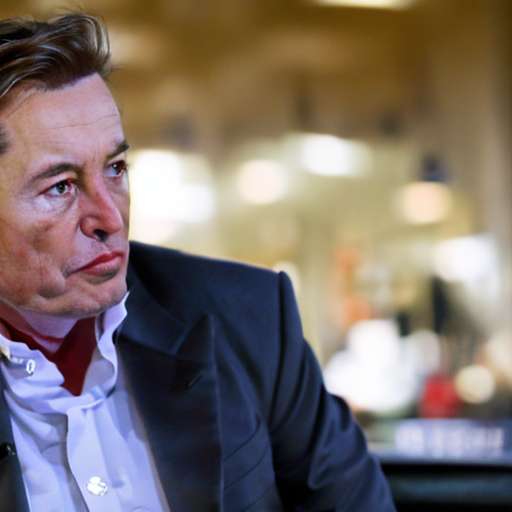}
		\caption{Elon Musk}
	\end{subfigure}
	\begin{subfigure}{0.24\linewidth}
		\includegraphics[width=\linewidth]{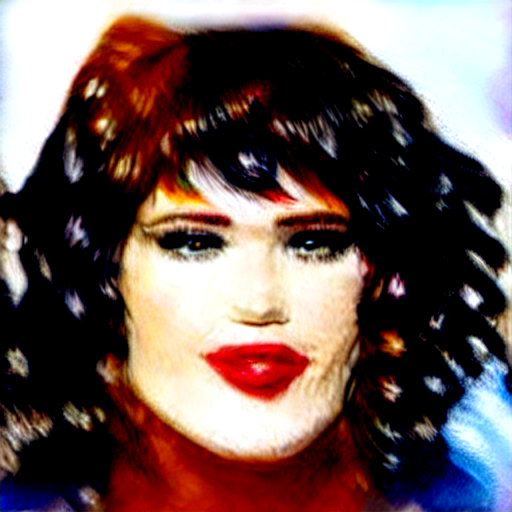}
		\caption{w/o}
	\end{subfigure}
	\begin{subfigure}{0.24\linewidth}
		\includegraphics[width=\linewidth]{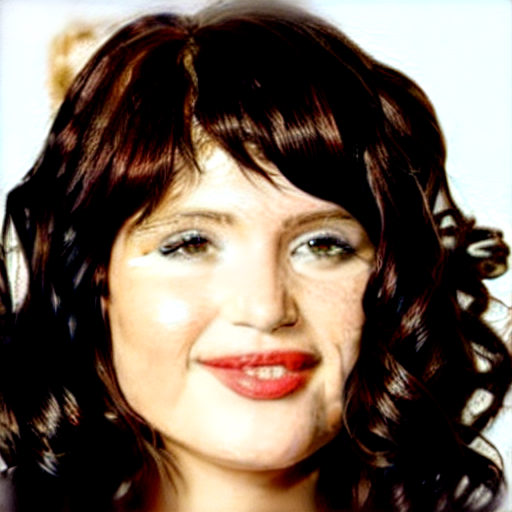}
		\caption{w/ }
	\end{subfigure}
	\begin{subfigure}{0.24\linewidth}
		\includegraphics[width=\linewidth]{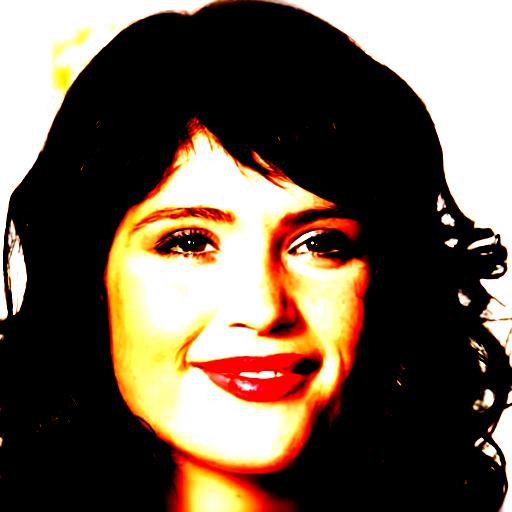}
		\caption{Private}
	\end{subfigure}
	\caption{The first image is the reconstruction result of Elon Musk by leveraging our attack not using variational encoder. The right three images are the reconstruction by our attack without and with variational encoder and an example of private image of a randomly picked identity of the validation set from CelebAHQ.}\label{fig:designVariationalEncoder}
\end{figure}

We let the final linear layer to generate a network embedding directly instead of mean and variance. The CLIP-T score is 23.93, CLIP-I score is 0.84, and recognition accuracy is 0.83. Example reconstruction and generation results are shown in \cref{fig:designVariationalEncoder}. Although using an autoencoder for a single data point produces images with identities similar to the original private images, there are minor differences, and performance is not as good as with a variational autoencoder. The reconstruction may overfit on specific features, making a variational autoencoder the better option.

\subsubsection{Ablation Study on Fitting Text Embedding Directly}\label{sec:fittext}
An interesting question arises in our autoencoder structure: why do we need to connect the network encoder to the fine-tuned diffusion model? In addition to the network encoder, we also connect the encoder to the fine-tuned diffusion model. Subsequently, when training the encoder, we utilize the fine-tuned diffusion model to output predictions of the noise and then propagate the loss backward to update the parameters of the autoencoder. Our objective is to find a suitable embedding for the fine-tuned diffusion model, enabling us to generate an image that reveals private information. A straightforward approach is to directly use the text embedding as the label and align the network embedding with the text embedding.

Therefore, we do an experiment in which we directly employ the mean square error loss (MSE) to align the output of the network encoder with the text embedding during the training process. The revised training procedure is illustrated in \cref{alg:fittext}, with the controlled variable being the update of the network encoder through the MSE between texts and network embedding. We anticipate that during the attacking phase, the network encoder can directly output the predicted text embedding, which we will then use to generate reconstructions of private images.

\begin{algorithm}[tb]
	\caption{The method of training network encoder fitting the network embedding to text embedding directly}\label{alg:fittext}
	\begin{algorithmic}[1]
		\STATE{{\bfseries Input:} The pre-trained diffusion model $\theta_0$ and a public dataset.}\;
		\STATE{{\bfseries Output: } The trained  network encoder with $M$ epochs.}\;
		\STATE{iteration $\leftarrow$ 0.}\;
		\WHILE{iteration $<M$}
		\STATE{Pick a batch of images from the public dataset: $\mathcal{X}$.}\;
		\STATE{Uniformly samplean $s'$ from $[1,s]$.}\;
		\FOR{fine-tuning rounds $r\leftarrow 1$ to $s'$}
		\STATE{iteration$\leftarrow$iteration$+1$.}\;
		\STATE{Generate Gaussian noise and add on to $\mathcal{X}$.}\;
		\STATE{Fine-tune diffusion model one step from $\theta_{r-1}$ to $\theta_r$.}\;
		\STATE{Model updates $\Delta \theta\leftarrow\theta_r-\theta_0$.}\;
		\STATE{Input $\Delta \theta$ and timestep $r$ into network encoder and output the network embedding.}\;
		\STATE{$\mathcal{Y}$ is the text embedding of the prompt paired to $\mathcal{X}$.}\;
		\STATE{Backward once on MSE loss between $\mathcal{Y}$ and network embedding.}\;
		\ENDFOR{}
		\ENDWHILE{}
	\end{algorithmic}
\end{algorithm}

\begin{figure}[tb]
	\centering
	\begin{subfigure}{0.24\linewidth}
		\includegraphics[width=\linewidth]{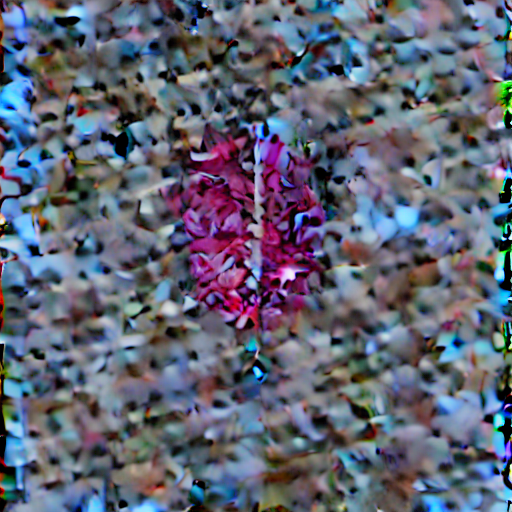}
		\caption{Elon Musk}
	\end{subfigure}
	\begin{subfigure}{0.24\linewidth}
		\includegraphics[width=\linewidth]{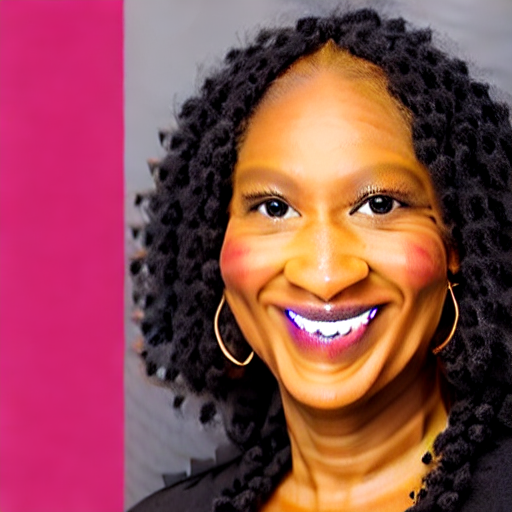}
		\caption{\cref{alg:fittext}}
	\end{subfigure}
	\begin{subfigure}{0.24\linewidth}
		\includegraphics[width=\linewidth]{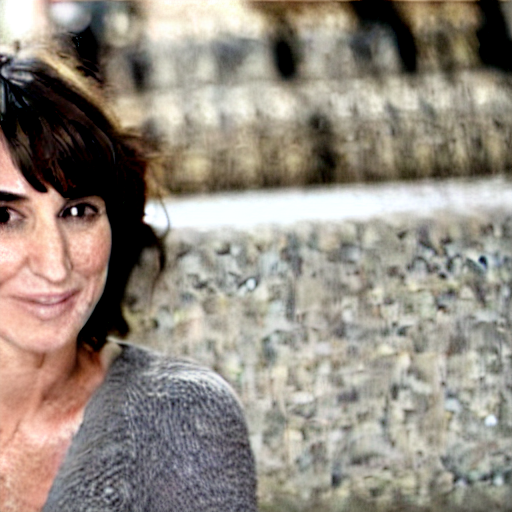}
		\caption{\cref{alg:efficient}}
	\end{subfigure}
	\begin{subfigure}{0.24\linewidth}
		\includegraphics[width=\linewidth]{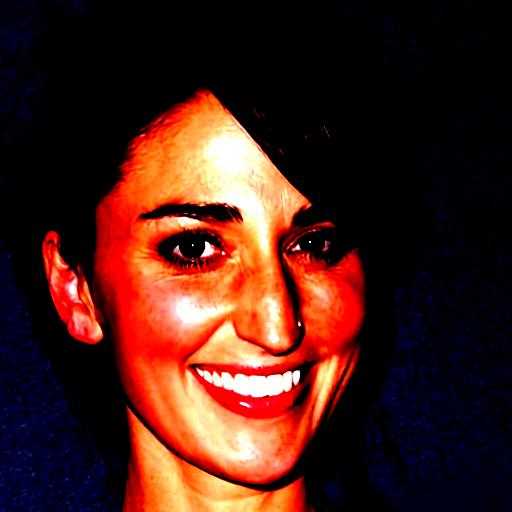}
		\caption{Private}
	\end{subfigure}
	\caption{The first image depicts the reconstruction result of Elon Musk using our attack-trained network encoder following \cref{alg:simpletrain}. The three images to the right showcase reconstructions obtained using our attack-trained network encoder following \cref{alg:simpletrain} and \cref{alg:efficient}, alongside an example of a private image featuring a randomly selected identity from the CelebAHQ validation set.}\label{fig:designFittext}
\end{figure}

However, on the CelebAHQ dataset during the attacking phase, the CLIP-T score is 24.01, the CLIP-I score is 0.75, and the recognition accuracy is 0.23. These similarity scores are far from the expected values. From \cref{fig:designFittext}, we observe that we are unable to generate the expected images depicting the private identity.

\begin{figure}[tb]
	\centering
	\includegraphics[width=\linewidth]{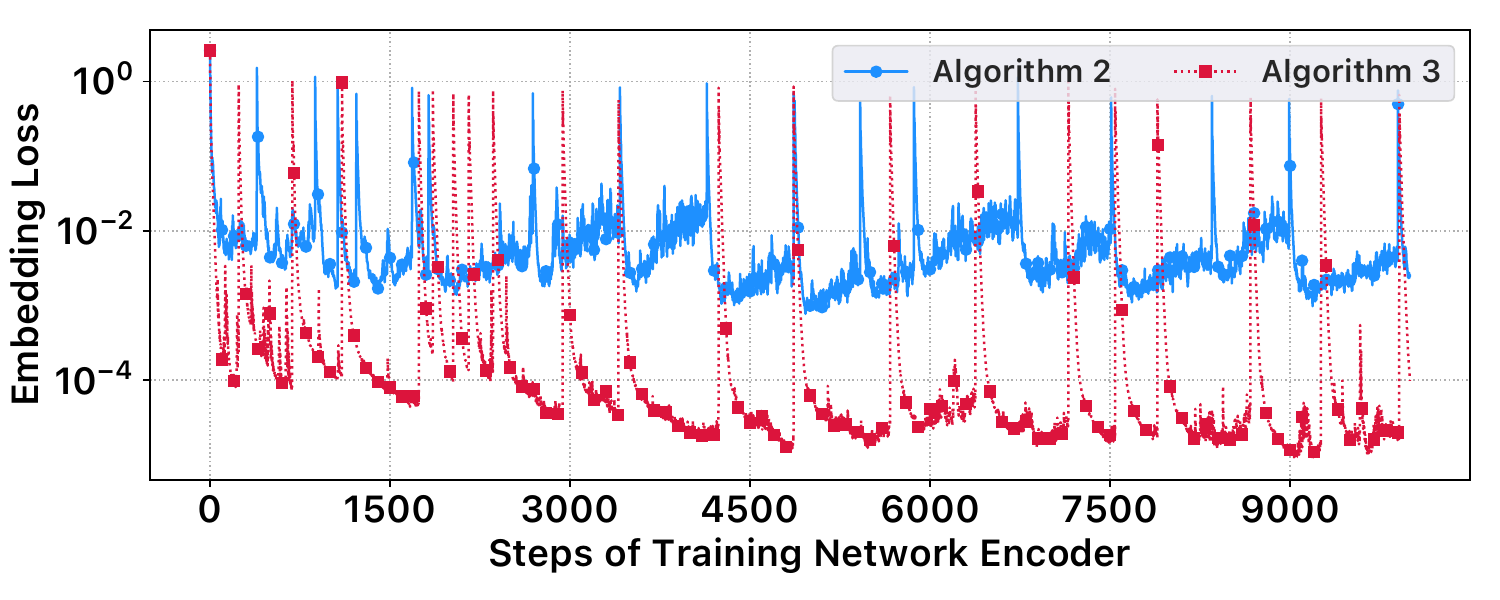}
	\caption{The changes in embedding loss during the first 10,000 steps to train the network encoder.}\label{fig:mseloss}
\end{figure}

To identify the proper reason, we examine the mean squared error (MSE) between the text and network embedding during the training process using different algorithms. We refer to this loss as the embedding loss. In the training process following \cref{alg:efficient}, we do not utilize this embedding loss to update the network encoder, but we still calculate the MSE between the text and network embedding. As depicted in \cref{fig:mseloss}, we observe that when using \cref{alg:fittext}, the loss quickly converges to a low value. However, during the attack, the network encoder fails to produce the expected embedding. This suggests that the network encoder is overfitting.

An interesting finding is that when training the network encoder following \cref{alg:efficient}, the embedding loss does not always converge. This shows our algorithm avoids overfitting. Additionally, reconstructing the text embedding is not always necessary to generate the proper image. Across the embedding space, multiple embeddings can trigger the fine-tuned diffusion model to generate an image with the expected private identity.

\section{Potential Defense}
To evaluate the effectiveness of our attack, we aim to explore defense methods against data reconstruction from diffusion model weights. In our threat model, the attacker only inputs model weights into the inversion network without interfering with fine-tuning.
\subsection{Experiment Settings}
\subsubsection{Defense Methods.} To defense against such an attack, the basic idea of existing work is to add distortion over the model weights so that an adversary cannot infer the deterministic information from model weights because of the introduced randomness. To achieve such an objective, the existing solutions are adding noise over the model weights following differential privacy. It is shown that among convex models and convolution neural networks for image classification, differential privacy~\cite{balle2022reconstructing} is the most effective defense against an adversary knowing model weights. 

The idea of a differential privacy method is straightforward, adding the Gaussian noise over the subject we want to protect, for example, model weights. Differential privacy establishes a rule to add such a noise for example to reach a certain privacy budget $\epsilon$, how large the magnitudes of the noise we should add~\cite{dwork2014algorithmic}. A formal definition of privacy budget $\epsilon$ is:
\begin{theorem}
 A randomized mechanism $\mathcal{M}$ mapping data from $\mathcal{X}$ on to $\mathcal{Y}$ is $(\epsilon,\delta)$ differential private if for any two datasets $x$ and $x'\in \mathcal{X}$ differ by at most one entry, and for any output sets $y\in\mathcal{Y}$, it holds that
 \[\Pr [\mathcal{M}(x)\in y]\leq e^{\epsilon} \Pr [\mathcal{M}(x')\in y]+\delta\]
\end{theorem}
According to this theorem, given the model weights after the differential privacy process is $y$, we cannot use it to reverse the original model weights $x$. To achieve the privacy budget of $(\epsilon,\delta)$, we need to add a Gaussian noise where the square of the variance should be bigger than $2\log(\frac{1.25}{\delta})\frac{C^2}{\epsilon^2}$~\cite{dwork2014algorithmic}. $C$ is a clip constant.

DPDM~\cite{dockhorn2023differentially} and DPGM~\cite{jiang2024functional} are two methods leveraging the differential privacy to protect privacy over diffusion model. With their methods, they are able to add distortion over the model weights, which have the potential to defend against our purposed attack. 

\subsubsection{Experiment Settings.} We evaluate our attack using differential private gradient descent to fine-tune models. In our experiments, according to the definition of $(\epsilon,\delta)$ differential privacy, we adopt the same setting of $\delta$ as $1\times10^{-6}$ in DPDM and DPGM\@. We choose three different values of $\epsilon$ as $0.1$, $2$, and $10$, the same values as DPDM and DPGM, to see the boundary of these defense methods. The $\sigma$-list used in DPGM is $[1, 2, 4, 8, 16]$.

\subsubsection{Evaluation Metrics.} In this section, we use the same set of evaluation metrics. We expect the similarities of our attacks to be high, which means that our attack method is effective and our defense methods do not work. Or we expect the similarities between images generated by DreamBooth and private images to be low, which means the defense methods corrupt the utility of the origin diffusion model. With such defense methods, the fine-tuned diffusion model cannot generate expected images correctly. If the similarity scores of DreamBooth is high and of our attack is low, it means the defend methods are effective. 

\subsection{Evaluate Effectiveness of Defense}
\begin{table}[tb]
 \caption{Effectiveness of our attack against different defense methods on CelebAHQ.}\label{tab:defense}
 \begin{center}
\begin{threeparttable}
 \begin{tabular}{ccccc}
 \toprule
 Defense&Source&CLIP-T&CLIP-I&Rec.~Acc.\\
 \midrule
 \multicolumn{5}{c}{$\epsilon=10$}\\
 \multirow{2}{*}{DPDM}& DreamBooth&26.24&0.73&0.96\\
 &Reconstruction&26.11&0.81&0.74\\
 \multirow{2}{*}{DPGM}& DreamBooth&24.48&0.91&0.98\\
 &Reconstruction&25.12&0.84&0.91\\
 \midrule
 \multicolumn{5}{c}{$\epsilon=2$}\\
 \multirow{2}{*}{DPDM}& DreamBooth&24.99&0.73&0.91\\
 &Reconstruction&26.65&0.75&0.031\\
 \multirow{2}{*}{DPGM}& DreamBooth&25.89&0.92&0.98\\
 &Reconstruction&25.71&0.91&0.91\\
 \midrule
 \multicolumn{5}{c}{$\epsilon=0.1$}\\
 \multirow{2}{*}{DPDM}& DreamBooth&26.25&0.75&0.88\\
 &Reconstruction&24.89&0.78&0.031\\
 \multirow{2}{*}{DPGM}& DreamBooth&24.26&0.79&0.97\\
 &Reconstruction&24.13&0.81&0.71\\
 \bottomrule
 \end{tabular}
\end{threeparttable}
 \end{center}
\end{table}

\begin{figure}[!htb]
	\centering
	\begin{subfigure}{\linewidth}
		\includegraphics[width=\linewidth]{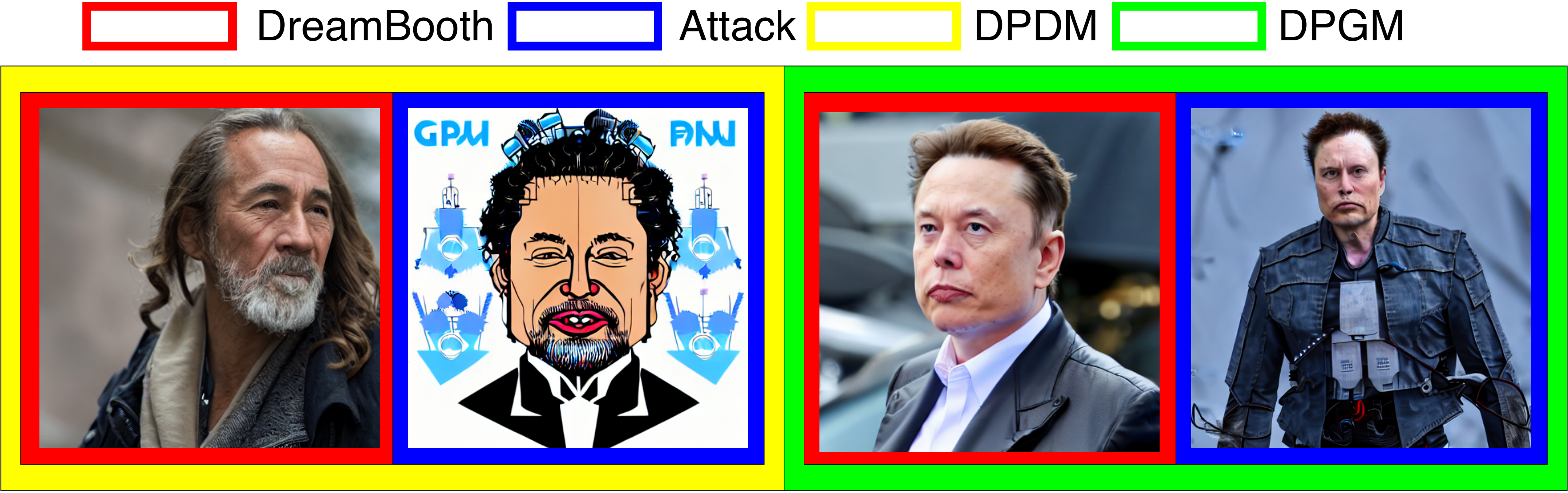}
		\caption{$\epsilon=10$}
	\end{subfigure}
	\begin{subfigure}{\linewidth}
		\includegraphics[width=\linewidth]{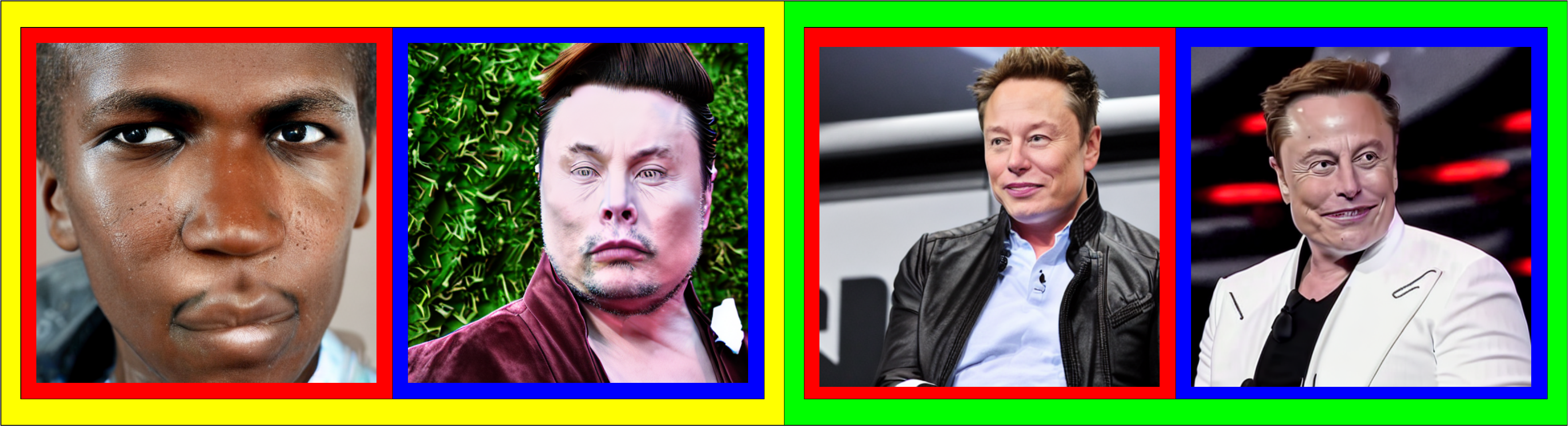}
		\caption{$\epsilon=2$}
	\end{subfigure}
	\begin{subfigure}{\linewidth}
		\includegraphics[width=\linewidth]{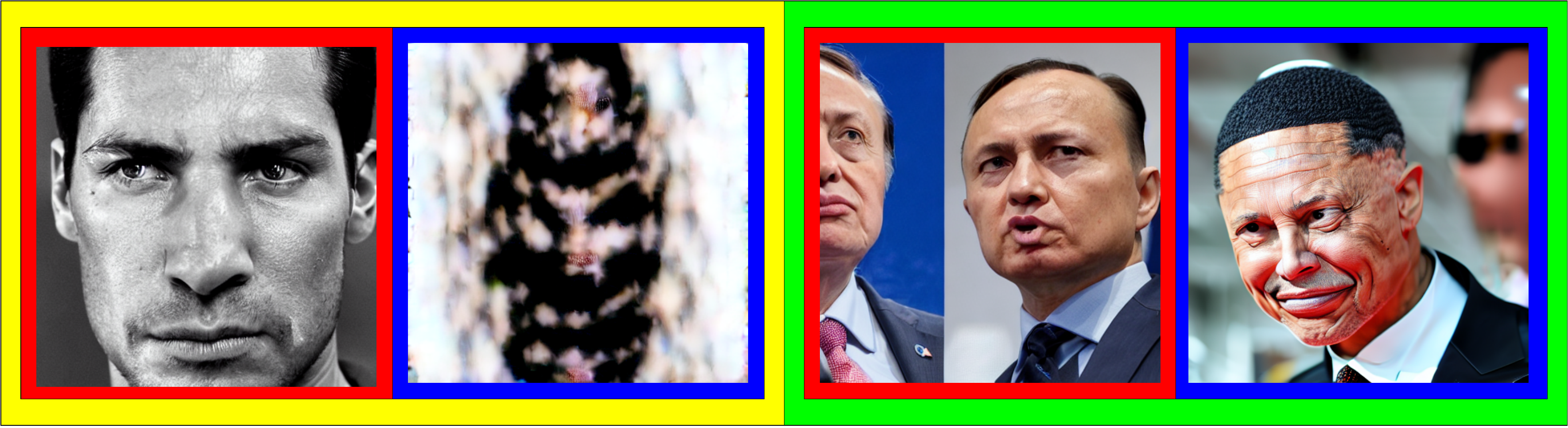}
		\caption{$\epsilon=0.1$}
	\end{subfigure}
	\caption{The images are generated by a Stable Diffusion V-1.4 model, fine-tuned with images of Elon Musk and tested against various defense methods using different values of $\epsilon$. From left to right, the images represent the output from the fine-tuned model with DPDM, the reconstruction against DPDM, the image generated by the fine-tuned model with DPGM, and the reconstruction against DPGM, respectively.}\label{fig:dpMusk}
\end{figure}

As shown in the \cref{tab:defense}, rows of DreamBooth are the similarities scores of the fine-tuned diffusion model. We can see that for the method DPDM, the utility of the model is corrupted. It cannot correctly generate images which are supposed to have higher similarity scores with the private images having the same identity. The scores are lower than those generated by DreamBooth in \cref{tab:attackcelebhq}. Also as depicted in \cref{fig:dpMusk}, we can see that models fine-tuned with DPDM fails to generate images having the faces of Elon Musk, even when $\epsilon=10$. Hence, DPDM is not an effective defense method as it preserves privacy at the cost of completely disrupting the utility of the diffusion model.

The other method is DPGM\@. When $\epsilon$ equals to $10$ and $2$, we can see that the fine-tuned diffusion model is able to generate the expected images. In \cref{fig:dpMusk}, we can see the faces of Elon Musk are shown in the generated by diffusion models. While in such a case, as the magnitudes of the Gaussian noise are not strong enough, our attack method is also able to reconstruct the private images with the model weights fine-tuned by DPGM\@. When we have a much smaller privacy budget where $\epsilon=0.1$, the magnitudes of the Gaussian noise is much larger. We can see that the utility of the fine-tuned diffusion model is disrupted. In such a case, we can see that our attack is successfully defended. However, we are not able to generate the images we want, having the faces of Elon Musk. We can see that, with such a defense method, we cannot preserve privacy without compromising the utility of the diffusion model.
\subsection{Further Discussion}
It is quite surpring that the method of adding noise, which is previously effective, does not work as expected on diffusion models. A reason is that in our attack method, we do not require to know the initial values of the LoRA matrices. The B matrices of LoRA matrices are initialized as 0 and use a random Gaussian initialization for A matrices~\cite{hu2021lora}. Though the model weights of these LoRA matrices are randomly initialized, our method still manages to reconstruct private images. Adding on Gaussian noise later is not so effective as these matrices are fine-tuned based on a randomly generated by Gaussian distribution. Such an attack is robust to random noise. As the training and attacking phases of our attack follow the similar steps in the training and inference phases in the diffusion process. The performance of the attack and generating images by fine-tuned models is coupled. In order to make such an attack ineffective, we need to add noises having relatively large magnitudes and such distortion is too large to maintain the utility of diffusion models. Hence, to defend against such a type of attack, we need to jump out of the box of adding noise and differential privacy and seek for other defense methods as a future work.

\section{Concluding the Remarks}
In this paper, we attempt to answer the question: given access to the fine-tuned weights of a diffusion model, can an adversary reconstruct the private data used for fine-tuning? Unlike previous literature, we establish a more practical assumption: that an adversary is unable to obtain prompts used for training, but only the fine-tuned model weights. To execute the attack, we design a variational network autoencoder composed of a specially designed network encoder to encode LoRA matrices into network embeddings and the fine-tuned diffusion model. During the attack phase, such an autoencoder can take the fine-tuned model weights as input and generate an image, serving as the reconstruction of private images. To enhance the efficiency of training such an encoder, we propose an efficient training paradigm with the help of timestep embedding. It is verified that such an attack can indeed reconstruct private images. Building on this attack, we explore a defense method based on adding Gaussian noise and employing differential privacy. We demonstrate that no existing privacy protection method can effectively defend against such an attack without compromising the utility of the fine-tuned diffusion model. This work encourages a second thought before sharing the weights of a fine-tuned diffusion model and leaves the defense method as an open question.

\section{Broader Impact}
In this paper, we aim to draw attention to the potential for leakage during the use of diffusion models. The possible implications of this work include further developing solutions to mitigate such privacy risks. Another impact will be raising people's attention to similar vulnerabilities such as the potential privacy leakage when fine-tuning a diffusion model with full-parameter fine-tuning.

\newpage
\bibliographystyle{IEEEtranN}
\bibliography{main}

\end{document}